\documentclass[12pt,twoside]{report}
\newcommand{\forceindent}{\leavevmode{\parindent=2em\indent}}
\usepackage[numbers]{natbib}
\usepackage{graphicx}
\usepackage{pdfpages}
\usepackage[export]{adjustbox}
\usepackage{float}
\usepackage{afterpage}
\usepackage{amsmath,amssymb}
\usepackage{multirow}
\usepackage[toc,page]{appendix}
\newcommand\blankpage{%
    \null
    \thispagestyle{empty}%
    \addtocounter{page}{-1}%
    \newpage}

\newcommand{\reporttitle}{Aff-wild Database and Aff-wild Net}
\newcommand{\reportauthor}{Alvertos Benroumpi}
\newcommand{\supervisor}{Dimitrios Kollias}
\newcommand{\degreetype}{Computing Science / Machine Learning}

\usepackage[a4paper,hmargin=2.8cm,vmargin=2.0cm,includeheadfoot]{geometry}
\usepackage{textpos}
\usepackage{natbib} 
\usepackage{tabularx,longtable,multirow,subfigure,caption}
\usepackage{fncylab} 
\usepackage{fancyhdr} 
\usepackage{url} 
\usepackage[english]{babel}
\usepackage{amsmath}
\usepackage{graphicx}
\usepackage{dsfont}
\usepackage{epstopdf} 
\usepackage{backref} 
\usepackage{array}
\usepackage{latexsym}
\usepackage[pdftex,pagebackref,hypertexnames=false,colorlinks]{hyperref} 

\hypersetup{pdftitle={},
  pdfsubject={}, 
  pdfauthor={},
  pdfkeywords={}, 
  pdfstartview=FitH,
  pdfpagemode={UseOutlines},
  bookmarksnumbered=true, bookmarksopen=true, colorlinks,
    citecolor=black,%
    filecolor=black,%
    linkcolor=black,%
    urlcolor=black}

\usepackage[all]{hypcap}

\def\@makechapterhead#1{%
  \vspace*{10\p@}%
  {\parindent \z@ \raggedright \sffamily
    \interlinepenalty\@M
    \Huge\bfseries \thechapter \space\space #1\par\nobreak
    \vskip 30\p@
  }}

\def\@makeschapterhead#1{%
  \vspace*{10\p@}%
  {\parindent \z@ \raggedright
    \sffamily
    \interlinepenalty\@M
    \Huge \bfseries  #1\par\nobreak
    \vskip 30\p@
  }}

\allowdisplaybreaks

\date{September 2018}

\begin{document}

\begin{titlepage}

\newcommand{\HRule}{\rule{\linewidth}{0.5mm}} 


\includegraphics[width = 4cm]{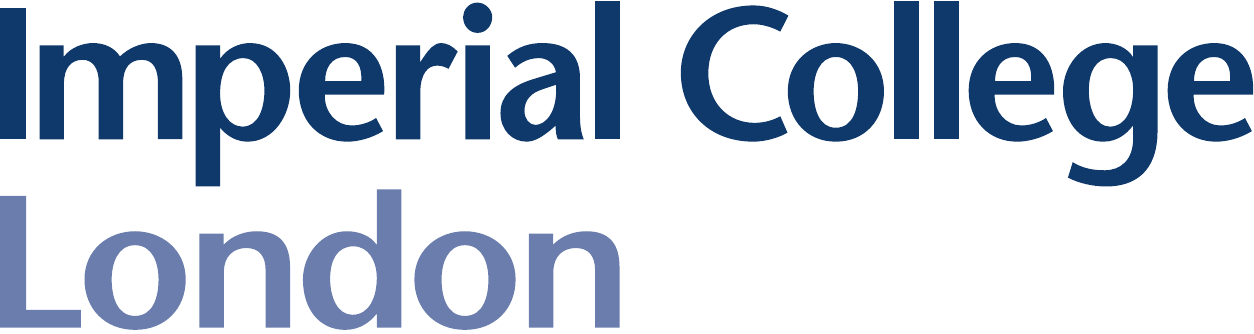}\\[0.5cm] 

\center 


\textsc{\Large Imperial College London}\\[0.5cm] 
\textsc{\large Department of Computing}\\[0.5cm] 


\HRule \\[0.4cm]
{ \huge \bfseries \reporttitle}\\ 
\HRule \\[1.5cm]
 

\begin{minipage}{0.4\textwidth}
\begin{flushleft} \large
\emph{Author:}\\
\reportauthor 
\end{flushleft}
\end{minipage}
~
\begin{minipage}{0.4\textwidth}
\begin{flushright} \large
\emph{Supervisor:} \\
\supervisor 
\end{flushright}
\end{minipage}\\[4cm]

\vfill 
Submitted in partial fulfillment of the requirements for the MSc degree in
\degreetype~of Imperial College London\\[0.5cm]

\makeatletter
\@date 
\makeatother

\end{titlepage}

\pagenumbering{roman}
\clearpage{\pagestyle{empty}\cleardoublepage}
\setcounter{page}{1}
\pagestyle{fancy}

\begin{abstract}
Emotions recognition is the task of recognizing people's emotions. Usually it is achieved by analyzing expression of peoples faces. There are two ways for representing emotions: The categorical approach and the dimensional approach by using valence and arousal values. Valence shows how negative or positive an emotion is and arousal shows how much it is activated. Recent deep learning models, that have to do with emotions recognition, are using the second approach, valence and arousal. Moreover, a more interesting concept, which is useful in real life is the "in the wild" emotions recognition. "In the wild" means that the images analyzed for the recognition task, come from from real life sources(online videos, online photos, etc.) and not from staged experiments. So, they introduce unpredictable situations in the images, that have to be modeled.\\
\\
The purpose of this project is to study the previous work that was done for the "in the wild" emotions recognition concept, design a new dataset which has as a standard the "Aff-wild" database, implement new deep learning models and evaluate the results. First, already existing databases and deep learning models are presented. Then, inspired by them a new database is created which includes 507.208 frames in total from 106 videos, which were gathered from online sources. Then, the data are tested in a CNN model based on CNN-M architecture, in order to be sure about their usability. Next, the main model of this project is implemented. That is a Regression GAN which can execute unsupervised and supervised learning at the same time. More specifically, it keeps the main functionality of GANs, which is to produce fake images that look as good as the real ones, while it can also predict valence and arousal values for both real and fake images. Finally, the database created earlier is applied to this model and the results are presented and evaluated.
\end{abstract}

\cleardoublepage

\fancyhead[RE,LO]{\sffamily {Table of Contents}}
\tableofcontents

\clearpage{\pagestyle{empty}\cleardoublepage}
\pagenumbering{arabic}
\setcounter{page}{1}
\fancyhead[LE,RO]{\slshape \rightmark}
\fancyhead[LO,RE]{\slshape \leftmark}
\afterpage{\blankpage}

\chapter{Introduction}
\section{Motivation}
\forceindent The past twenty years, the research in facial behavior analysis was limited to data that were specifically created for this scope, with experiments that contained data recorded in highly controlled situations. Moreover, most of these experiments which are trying to represent humans emotions, are based on categorical emotion representation, which includes the seven basic categories, i.e. Anger, Disgust, Fear, Happiness, Sadness, Surprise and Neutral. However, the dimensional emotion representation is more appropriate to represent the whole range of humans emotions. The most usual dimensional emotion representation, which is used on this project, is the 2-D Valence and Arousal space. In order to make good use of this emotional range, various facial expressions have to be analyzed from videos "in the wild". More specifically, these will be videos, not made specifically for emotion analysis, which include people's faces reacting to something.
\section{Aims}
\forceindent For the first part of the project, the aim is to create a new "in the wild" dataset, using web shared videos as source. The whole procedure of creating the dataset has to be followed, by using as guidance the \cite{kollias1} paper. That includes the pre-processing of the data, the face detection \cite{avrithis2000broadcast} and the annotation. Hopefully, all the range of emotions will be covered from this dataset in terms of valence and arousal. 
\forceindent Before moving to the second part of the project, previous work on applying this type of data to deep learning models will be studied. Moreover, state of the art networks such as GANs and Capsule networks will be also studied. Finally, it will be decided which model is going to be implemented with the new dataset and the results will be evaluated by comparing them with the results of previous works.

\section{Outcomes}
\forceindent The final dataset consists of 507.208 frames gathered from 106 videos and its size is approximately 48GB. These frames includes approximately 150 faces of different people. The distribution of valence and arousal values was good and it did not leave uncovered values. However, the annotation was carried by one person and that makes the annotations biased. Hence, the results of the models may be a little better than the normal.

\forceindent For the deep learning part of the project, it was decided to implement a standard CNN network just to test the data at first. However, The basic implementation of the project is a semi-supervised GAN that generate images and performs emotions prediction at the same time. Although that the results were acceptable, there are many improvements that can be made to the model, in order to perform better both in terms of quality of generated images and in the accuracy of the regression.


\chapter{Background and Related Work}
\section{Emotions Representation}
\subsection{Categorical Approach}
\forceindent The most common approach to classify emotions is the one proposed in \cite{categ}. According to P. Ekman, based on facial expressions, emotions can be categorized in as the following:\\
\begin{itemize}
  \item Anger
  \item Disgust
  \item Fear
  \item Happiness
  \item Sadness
  \item Surprise
\end{itemize}
\forceindent This way of representing emotions has been used in many relevant projects until now \cite{kollias6}. However, people's emotions is something complex and a classification with only six classes cannot be entirely realistic about the real emotion that a person expresses. A broader range of emotions would be preferable, in order for people's  emotions to be classified better.  
\subsection{Dimensional Approach}
\forceindent The first concept that was studied and is necessary for this project is Valence and Arousal. The most usual dimensional emotion representation is the 2-D Valence and Arousal space, which in contrary with categorical representation, it can represent a much bigger range of emotions. Valence and arousal are describing the reaction of a person in an event by taking values in a 2-D space. Valence ranges from highly positive to highly negative, whereas arousal ranges from calming or soothing to exciting or agitating. Hence, these two values are rating someone's reaction according to how positive or negative and how intense this reaction is. The graph below gives a better understanding of these two values \cite{VA2}:\\
\\
\begin{figure}[H]
  \centering
  \begin{minipage}[b]{1\textwidth}
    \includegraphics[scale=0.6, left]{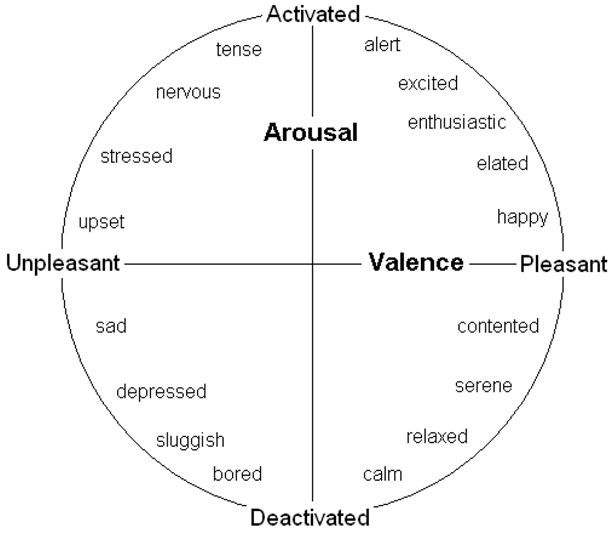}
    \caption{Example 2D dimensional space of valence-arousal. \\ source: https://www.pinterest.com/pin/354588170647498721/}
  \end{minipage}
\end{figure}
\forceindent If one thinks valence and arousal as a measure of someone's emotional condition, then it would be easy to think that these two values can be connected with one's face expressions. More specifically, the idea is that people tend to have similar face expressions when they want to express particular emotions. Hence, a pattern between this face-expression - emotion is created \cite{kollias10}, which it may can be decoded with a machine learning model \cite{kollias11,kollias12}, by having images as input and valence and arousal as output.

\section{Existing Databases}

\subsection{SEMAINE}
\forceindent In paper \cite{SEMAINE}, SEMAINE has created a large audiovisual database as part of an iterative approach to building agents that can engage a person in a sustained, emotionally colored conversation, using the Sensitive Artificial Listener (SAL) \cite{SAL} paradigm. The database consists of 150 participants who were undergraduate and postgraduate students, for a total of 959 conversations with individual SAL characters, lasting approximately 5 minutes each.  Videos are recorded at 49.979 frames per second and at a spatial resolution of 780 x 580 pixels.

\forceindent For the annotation part, SEMAINE members decided to include 5 dimensions which are well established in the psychological literature. These dimensions are  Valence, Activation, Power, Expectation and Intensity.  Valence and Activation(as Arousal) are described above, but the other dimensions are not introduced yet. The power dimension combines the power and control concepts of people and it depends on what they are facing. Anticipation/Expectation also subsumes various concepts that can be separated -– expecting, anticipating, being
taken unawares. Intensity, is about the distance of a person from a completely calm state.

\forceindent After the annotator had rated the data in terms of the five basic dimensions, he/she had to choose between 4 categories of labels to rate. These categories are Basic Emotions, Epistemic states, Interaction Process Analysis and Validity. Basic emotions are the following 7 emotions: Fear, Anger, Happiness, Sadness, Disgust, Contempt, Amusement. Epistemic states are states that can occur in people's interaction (e.g. Certain / not certain, Agreeing / not agreeing, Interested / not interested). Interaction Process Analysis includes behavior that people show in a dialogue (e.g.  Solidarity, Antagonism, Tension). Finally, Validity is a set of labels that shows if there is a contradiction between the feelings expressed from the participant and his/her real feelings.

\subsection{RECOLA}
\forceindent In paper \cite{Recola}, RECOLA is introduced, which stands for REmote COLlaborative and Affective interactions. For the construction of this database, participants were recorded in dyads during a video conference while completing a task requiring collaboration. Participants (mean age: 22 years, standard deviation: 3 years) were in total 46 (27 females, 19 males) and they were recruited as 23 dyadic teams work. 
All subjects are French speaking while having different mother tongues: 33 are originally French speaking, 8 Italian, 4 German and 1 Portuguese. According to the self-reports filled by the users, only 20\% of participants knew well their teammate. 

\forceindent This database is annotated in terms of valence and arousal. During the procedure, various techniques of mood induction and emotion manipulation were used, in order to have a broader range of annotated data. Annotators used the tool developed for RECOLA project, called ANNEMO. In order to be familiar with this tool before annotating the database, they first rated 2 sequences from SEMAINE database. 

\forceindent In order to keep and annotate only the interesting part of the recordings, only the first five minutes were kept for each one. Hence, from more than 9.5 hours of recordings, a reduced set of 3.8 hours was obtained.

\subsection{SEWA Database}
\forceindent The SEWA database \cite{sewa} is a set of visual and audio content, that was collected for the SEWA project. People participated in this project are from six different countries with a broad range of ages. At the registration of the project, a form with demographic measures was fulfilled by the participants. For the data collection, volunteers have to complete two experiments and they are separated based on their cultural background, age and gender.  

\forceindent For the first experiment, participants had to watch four adverts which were supposed to trigger emotions including amusement, empathy, liking and boredom. After watching the adverts, they had to complete a questionnaire about their real emotional state while watching the adverts.

\forceindent For the second experiment, volunteers had to do an online web-chat with another volunteer about the last advert they watched. More specifically, it was supposed to be a general discussion about their opinion for the advert and the product that was advertised. As before, after the discussion they had to complete a questionnaire about their emotional state during the conversation.

\forceindent The final database after the above experiments, contained 199 sessions of recordings, with a total of 2.075 minutes of recorded data, from 398 individual people.

\subsection{AFEW-VA Database}
\forceindent AFEW-VA database \cite{afew} contains data that are coming from films. More specifically 600 or more videos were extracted, with their length varying from short(10 frames) to long(more than 120 frames) videos. Moreover, the videos include scenes with challenging image conditions, which considering the background of the image or the people that are included.

\forceindent The way that the annotation procedure was carried out in AFEW-VA database, made the annotations to be very detailed and very accurate. That's because the annotators were 2 experts, who had to annotate the data per frame with values in the range of [-10, 10]. When the annotators did not agree about the values, they had a discussion about the content they were watching and finally they were coming up with a unique solution.
\subsection{The Aff-Wild Database}
\forceindent As mentioned above, except from the databases described before, there is one more existing database consisting of videos with more than 30 hours length. These videos were collected from YouTube and they were chose with various criteria. The common key-word that was used was the word "reaction". These videos consist of people reacting to different things, such as unexpected movie plots, activities or jokes \cite{kollias1,kollias2,kollias3}. This database has been already annotated in terms of valence and arousal, which are the two dimensions described before. The aim is to find more similar videos and enrich the already existing database by covering the whole range of valence and arousal values, in order to improve the performance of the predictive model.\\
\begin{figure}[H]
  \centering
  \begin{minipage}[b]{1\textwidth}
    \includegraphics[scale=0.28, left]{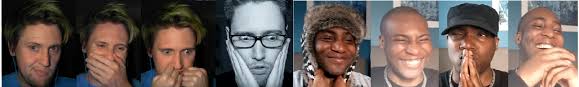}
    \caption{Some representative frames from aff-wild database.\\ source: \cite{kollias1,kollias2,kollias3}}
  \end{minipage}
\end{figure}

\subsubsection{Annotation of Aff-Wild Database}
\forceindent For the annotation part, a software has already been implemented. This software gives the ability to continuously annotate a video, as time passes, for valence and arousal individually, which is less demanding for the annotator, so the annotation quality may be higher. The annotator follows the annotation instructions and uses a joystick in order to perform the labeling. The instructions include examples of well identified frames with their valence and arousal values. Moreover, in the existing database, some data pre-processing and post-processing procedures have been implemented, which are analytically described in \cite{kollias1,kollias2,kollias3}.
\section{Neural Networks}
\subsection{CNN (Convolutional Neural Network)}
\forceindent Convolution Neural Networks, know as CNN, are widely used for image processing tasks after 2012 \cite{kollias13,simou2007fire,tagaris1,simou2008image,kollia2010semantic,tagaris2}. Big companies such as Facebook, Amazon and Google \cite{horrocks2011answering,glimm2013using} are using CNNs for various services that require deep learning solutions. CNNs are going to be used in this project because the videos in our database consist of frames, namely images.

\forceindent An image consists of pixels, so an image-input is basically an array of pixels. If we had an image 320x320, then we would have an array 320x320x3(in RGB format) as input. An overview of the structure of CNNs would be that we give the input, CNN pass it through a series of convolutional, nonlinear, pooling(downsampling), and fully connected layers and finally we get an output \cite{CNN}. A high level description for the convolutional layers would be that they work as feature identifiers. For every layer there is a filter of a particular size that scans the whole image and returns an activation map. By creating a series of convolutional, ReLU or pool layers we basically detect high level features of the image such as curves, edges, etc. The output(activation map) of these layers is the input for a fully connected layer, which returns the final output of our model, as usual.\\
\begin{figure}[H]
  \centering
  \begin{minipage}[b]{1\textwidth}
    \includegraphics[scale=0.37, left]{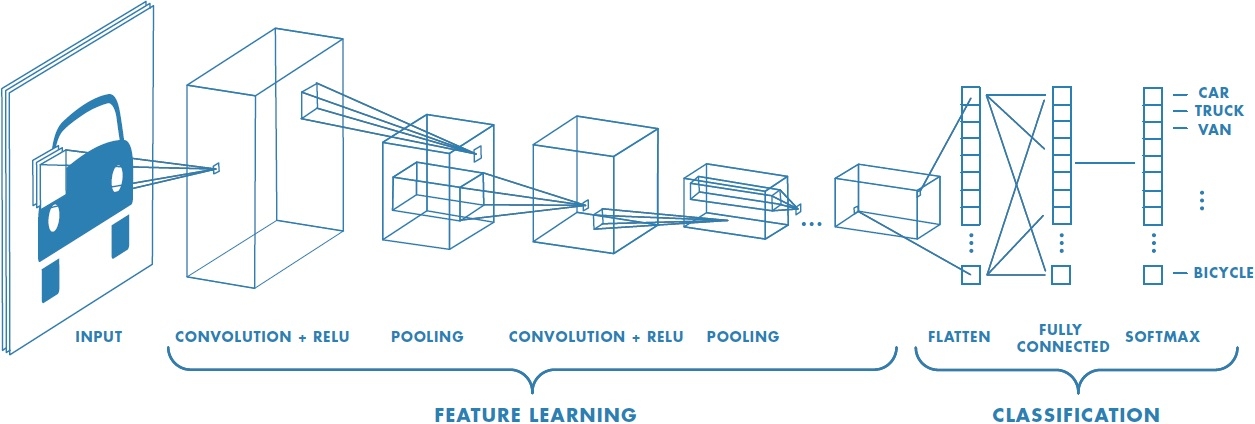}
    \caption{How a CNN works. \\ source: https://www.mathworks.com/discovery/convolutional-neural-network.html}
\end{minipage}
\end{figure}

\forceindent More specifically, when a new image is presented to the network, it is unknown where the features are. This is where convolution layers are introduced. For each feature, a filter is created which scans the whole image and it creates a map which indicates in which area of the image the feature has a strong appearance. This is achieved by applying convolutions between the filter and the image areas. Convolution layers result to a set of filtered images, that will be fed to the next layers. 

\forceindent Another common tool that is used in CNNs is pooling layers. For every image in the set of filtered images that resulted above, the pooling layer is reducing its size. It is actually a small window that scans the image with a specific step and for the set of pixels that are inside the window, it returns the maximum pixel value (maxpooling). It finally returns a set of images, with the same number of images as before, but their size is smaller.

\forceindent The activation function that is commonly used between a set of convolution-pooling layers is called ReLU \cite{relu}. ReLU is a very simple mathematical operation that substitutes every pixel of the image which has a negative value with zero. This helps the CNN stay mathematically healthy by keeping learned values from getting stuck near 0 or blowing up toward infinity.

\forceindent Either for a regression or classification task, the last step of the CNNS are the fully connected layers. These layers treats the features input as a list and they are matching each feature with its most probable class. Their output size is the number of features or classes that they have to predict.

\begin{figure}[H]
  \centering
  \begin{minipage}[b]{0.45\textwidth}
    \includegraphics[scale=0.42, left]{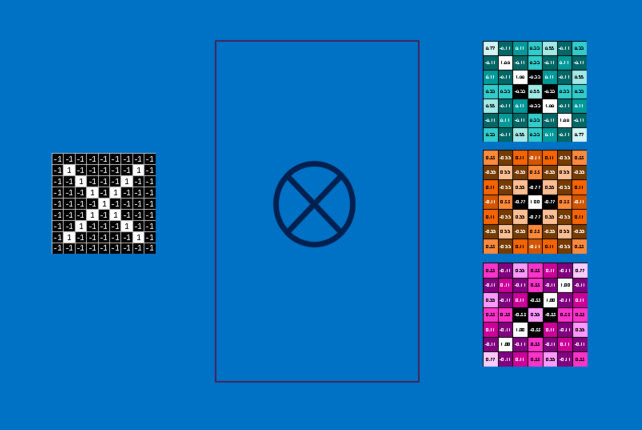}
    \caption{Result after convolution layer.}
  \end{minipage}
  \hfill
  \begin{minipage}[b]{0.45\textwidth}
    \includegraphics[scale=0.4, right]{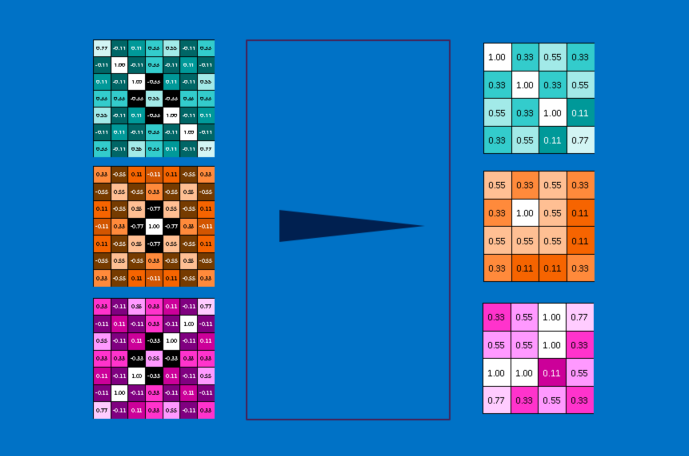}
    \caption{Result after maxpooling layer.}
  \end{minipage}
\end{figure}

\begin{figure}[H]
  \centering
  \begin{minipage}[b]{0.45\textwidth}
    \includegraphics[scale=0.45, left]{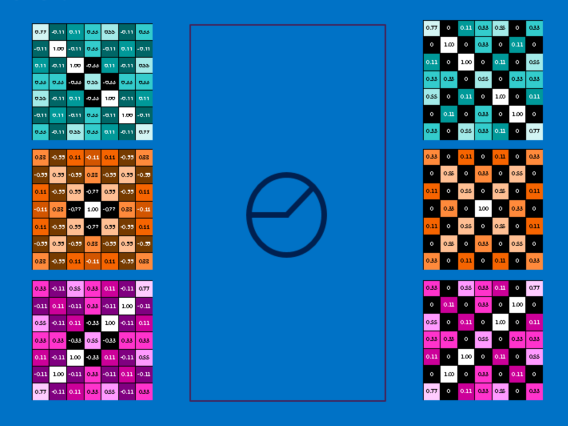}
    \caption{Result after ReLU activation.}
  \end{minipage}
  \hfill
  \begin{minipage}[b]{0.45\textwidth}
    \includegraphics[scale=0.42, right]{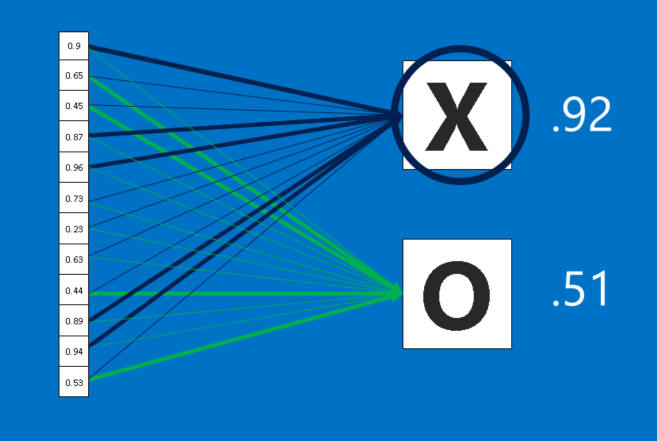}
    \caption{Result after fully connected layer (for images of X and O classification).}
  \end{minipage}
\end{figure}
source: https://brohrer.github.io/how\_convolutional\_neural\_networks\_work.html

\subsection{RNN (Recurrent Neural Network)}
\forceindent As described in \cite{RNN_LSTM}, Recurrent Neural Networks (RNN) are networks with loops in them, allowing information to persist. RNNs share parameters across different positions/ index of time/ time steps of the sequence, which makes it possible to generalize well to examples of different sequence length. RNN is usually a better alternative to position-independent classifiers and sequential models that treat each position differently.

\forceindent In the last few years, they are broadly used in projects, such as speech recognition and language modeling, which include sequential data and time-series. For this project, the ability of RNNs to keep information may be useful by using previous video frames to help the understanding of a present frame.\\
\begin{figure}[H]
  \centering
  \begin{minipage}[b]{1\textwidth}
    \includegraphics[scale=0.85, left]{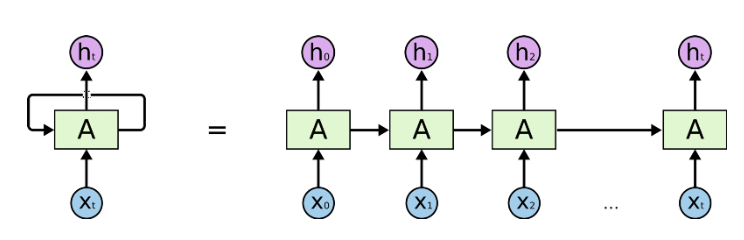}
    \caption{An unrolled recurrent neural network.\\ source: http://colah.github.io/posts/2015-08-Understanding-LSTMs/}
  \end{minipage}
\end{figure}

\subsubsection{LSTM (Long Short Term Memory)}
\forceindent For many tasks like these there is also a special kind of RNN , the LSTM. LSTMs are explicitly designed to avoid the long-term dependency problem. As it is analytically described in \cite{RNN_LSTM}, they work as filters of the previous state of the NN, which have the ability to remove or add information to the cell state, carefully regulated by structures called gates. LSTMs are very useful and they have contributed in the success of RNNs, because they can keep the information that passes from state to state clean, even for a long sequence of data.\\
\\
\begin{figure}[H]
  \centering
  \begin{minipage}[b]{1\textwidth}
    \includegraphics[scale=0.5, left]{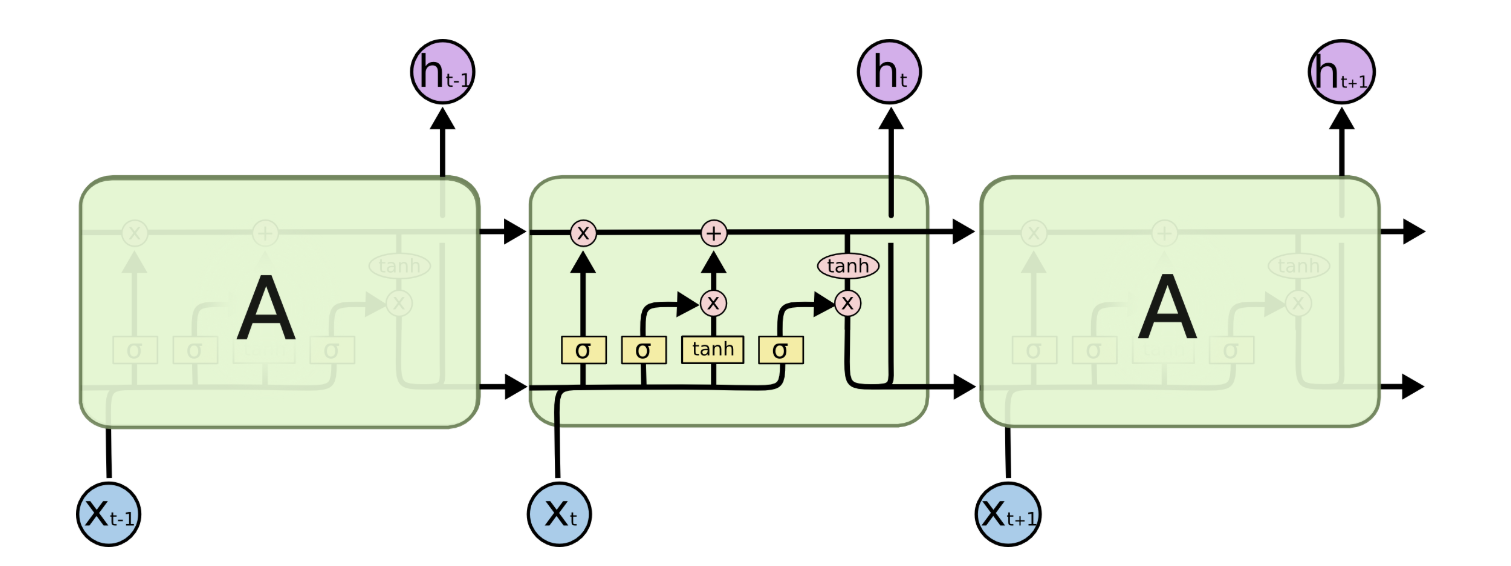}
    \caption{The structure of a LSTM cell. \\ source:  http://colah.github.io/posts/2015-08-Understanding-LSTMs/}
  \end{minipage}
\end{figure}

\subsubsection{GRU (Gated Recurrent Unit)}
\forceindent GRU can be considered as a variation on the LSTM because they have similar design and, in some cases, produce equally excellent results. In order to solve the vanishing gradient problem of a standard RNN, GRU uses two gates, the update and the reset gate. By using these gates, GRUs are able to store and filter information. Hence, the vanishing gradient problem is not present anymore, because the model is not washing out the new input every single time but keeps the relevant information and passes it down to the next time steps of the network. If GRUs get trained carefully, they can have an outstanding performance in difficult scenarios. The mathematics behind GRUs are explained in detail in \cite{GRU}.\\
\\
\begin{figure}[H]
  \centering
  \begin{minipage}[b]{1\textwidth}
    \includegraphics[scale=0.5, left]{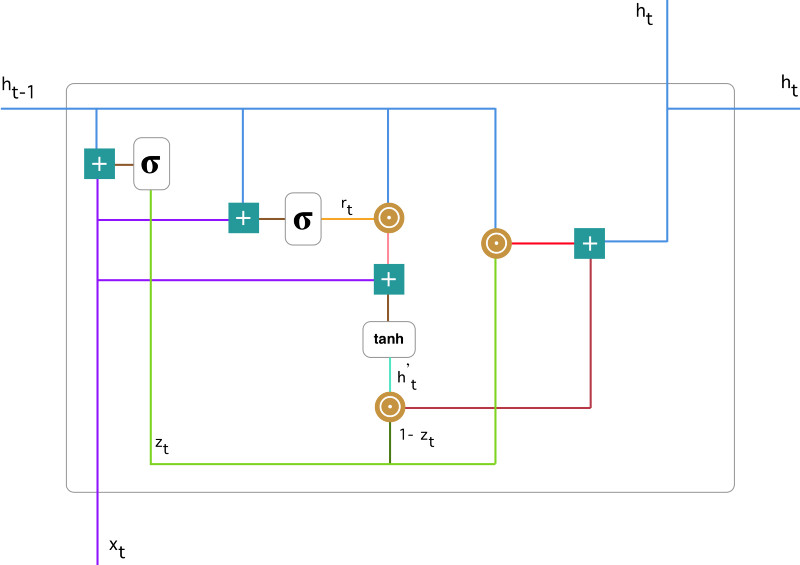}
    \caption{The structure of a GRU cell. \\ source:  https://towardsdatascience.com/understanding-gru-networks-2ef37df6c9be}
  \end{minipage}
\end{figure}

\subsection{GANs (Generative adversarial networks)}
\forceindent In 2014, a new framework for estimating generative models via an adversarial process was proposed \cite{GAN}. This framework includes a generative model G and a discriminative model D. The generative model takes as input a random noise and produces a sample (e.g. an image). The discriminative model takes as input samples(training data), which are same type as the output of G, and gives as an output the probability that the sample from G came from the training data rather than G. \\
\\
\begin{figure}[H]
  \centering
  \begin{minipage}[b]{1\textwidth}
    \includegraphics[scale=0.5, left]{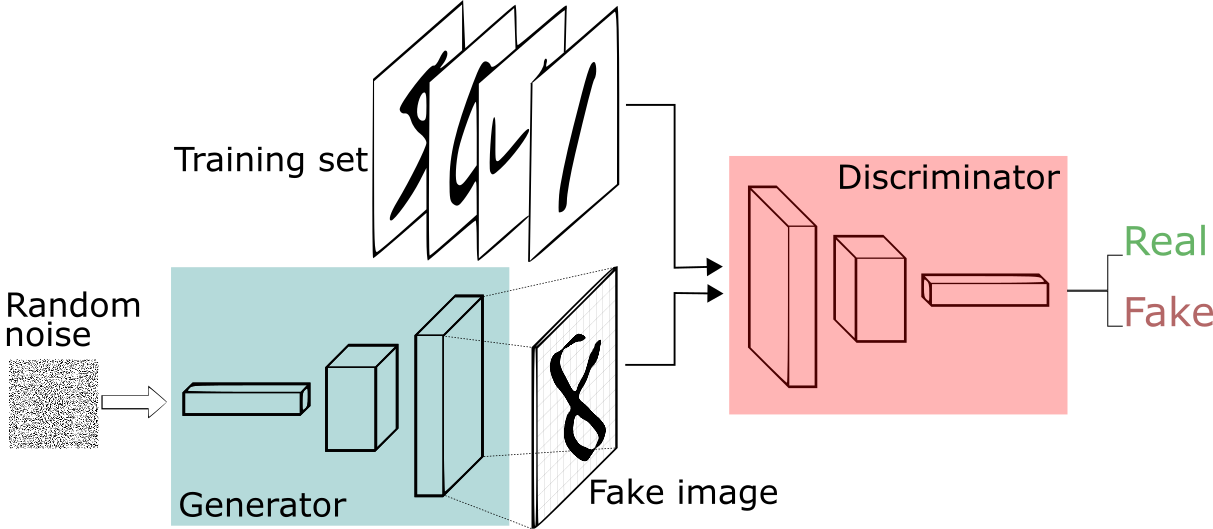}
    \caption{The overview of a GAN. \\ source:  image source: https://deeplearning4j.org/generative-adversarial-network}
  \end{minipage}
\end{figure}

\forceindent The framework applies better when the two models are both multilayer perceptrons. G is  differentiable function represented by a multilayer perceptron with parameters $\theta_g$ and a mapping to data space is represented by $G(z;\theta_g)$, where $p_z(z)$ is  a prior on input noise variables. $D(x;\theta_d)$ is a second multilayer perceptron that outputs a scalar, where x is the input data. The output of D is the probability that x came from the data rather than $p_g$. The goal is to train D to maximize the probability of assigning the correct label to both the samples of G(fake) and data from x(real). Accordingly, G have to be trained to minimize $\log(1-D(G(z)))$. Hence, the following minimax game is created by G and D with value function V(G,D):\\
\\
$\min_G \max_D V(D,G) = E_{x\sim p_{data}(x)}\lbrack \log(D(x))\rbrack+ E_{z\sim p_z(Z)}\lbrack \log(1 - D(g(Z)))\rbrack$.
\\

\forceindent One application of GANs that could be proved useful for this project is data augmentation \cite{kollias8,caridakis2006synthesizing,kollias9}. If a GAN is fed with frames for a particular emotional state (or in our case a particular valence-arousal range), then it could produce more frames similar with these we used as input, which express the same emotional range. With this technique, the current database could be enriched with data and a better performance could be achieved for the main emotions classification model.

\subsection{DCGAN}
\forceindent According to the \cite{DCGAN} paper, historical attempts to scale up GANs using commonly accepted and efficient CNN architectures have been unsuccessful. But by applying some modifications on a typical CNN architecture, a more stable training process across a range of databases was achieved. Those modifications are targeting three parts of a CNN architecture. 

\forceindent First of all, the paper proposes to substitute all pooling function (such as maxpooling) with strided convolutions, in order to allow network to learn from its own spatial downsampling. This modification can be used in both the generator and the discriminator.

\forceindent The second change concerns the connection between the highest convolution layer and the input or output of the generator or the discriminator respectively. Eliminating fully connected layers and using global average pooling increased the stability of the model but hurt convergence speed. So, a middle ground of these two is to directly connect input/output with convolution layers by using just a reshape. 

\forceindent Last modification that was implemented, is the use of batch normalization \cite{BN}, which stabilizes the input by assigning zero mean of variance of one. Batch normalization solves problems that can occur from bad initialization and helps gradient flow in deeper models. Moreover, it provided great contribution when used in generator, preventing it from collapsing all samples to a single point, which is a common problem that occurs in GANs. However, the application of batch normalization to the last layer of generator and the first layer of discriminator was avoided, because it resulted to model instability.

\forceindent Another point that is made in this paper, is that it was observed that using ReLU \cite{relu} activation function for the generator, except for the output layer that tanh was used, and leaky ReLU \cite{lrelu} for the discriminator was very beneficial for the training process.

\subsection{SSGAN}
\forceindent Chapter five of \cite{SSGAN} introduces Semi-Supervised learning in GANs. More specifically, it presents a classic classifier which classifies a data point x into K possible classes, which is trained like a standard supervised classifier. So, the paper proposes to convert the learning to semi-supervised by simply adding the generated samples to the dataset and labeling them with a new class y=K+1, so the classifier output dimension while be K+1. After this change, the classifier could learn from unlabeled data, as long as these data correspond to one of the K classes of real data.

\forceindent Supposing the half of the data are real and half generated the Loss function for the classifier/discriminator will be:\\
\[L=L_{supervised}+L_{unsupervised},\quad where\]\\
\[L_{supervised}=-\mathbb{E}_{x,y\sim p_{data}(x,y)}\log p_{model}(y|x,y<K+1)\]
and
\[L_{unsupervised}=-\{\mathbb{E}_{x\sim p_{data}(x)} \log \lbrack 1-p_{model}(y=K+1|x) \rbrack + \mathbb{E}_{x\sim G} \log \lbrack p_{model}(y=K+1|x) \rbrack \} \]

\begin{figure}[H]
  \centering
  \begin{minipage}[b]{1\textwidth}
    \includegraphics[scale=0.22, left]{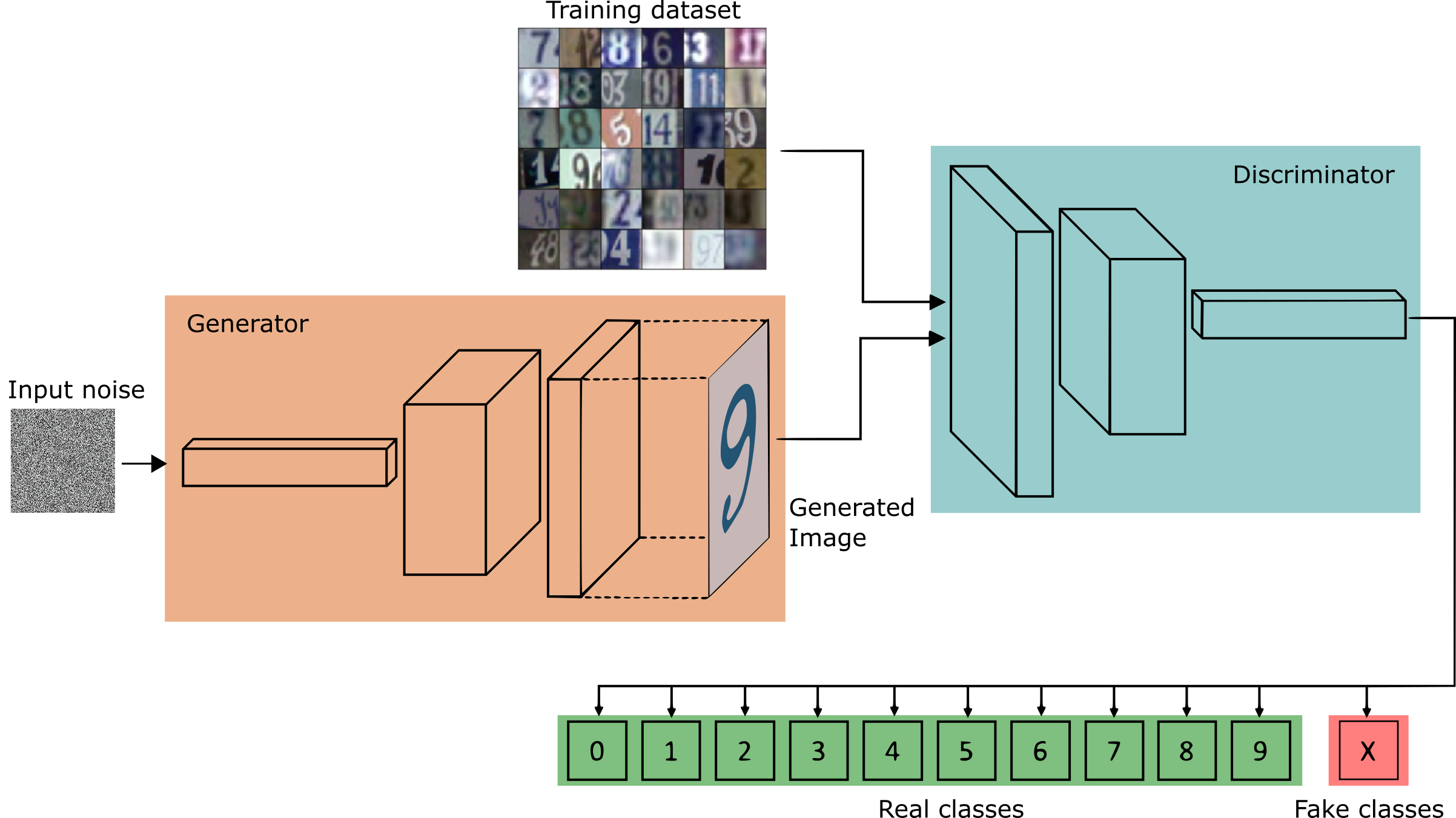}
    \caption{SSGAN architecture.\\
    source: https://sthalles.github.io/assets/semi-supervised/GAN\_classifier.png}
  \end{minipage}
\end{figure}

\forceindent SSGAN is the closest implementation to the network that is going to be created later for this project \cite{kollias5}. The difference is that the classification problem has to be converted in regression problem. That affects the output of the discriminator and also both the unsupervised and supervised loss for the discriminator. The results of the SSGAN for the cifar-10 \cite{cifar} will be presented below, because they will be useful to compare with the Regression GAN model of this project.

\begin{figure}[H]
  \centering
  \begin{minipage}[b]{1\textwidth}
    \includegraphics[scale=0.6, left]{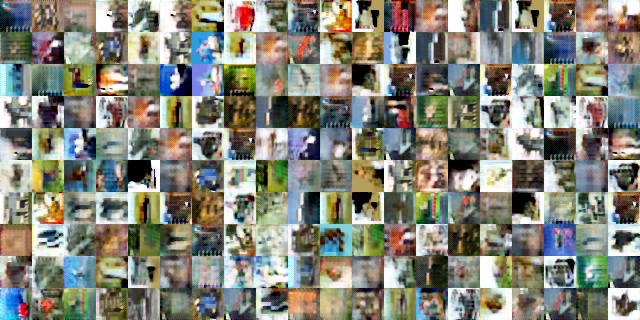}
    \caption{Generated samples (1000th epochs).\\
    source: https://github.com/gitlimlab/SSGAN-Tensorflow}
  \end{minipage}
\end{figure}

\begin{figure}[H]
  \centering
  \begin{minipage}[b]{1\textwidth}
    \includegraphics[scale=0.32, left]{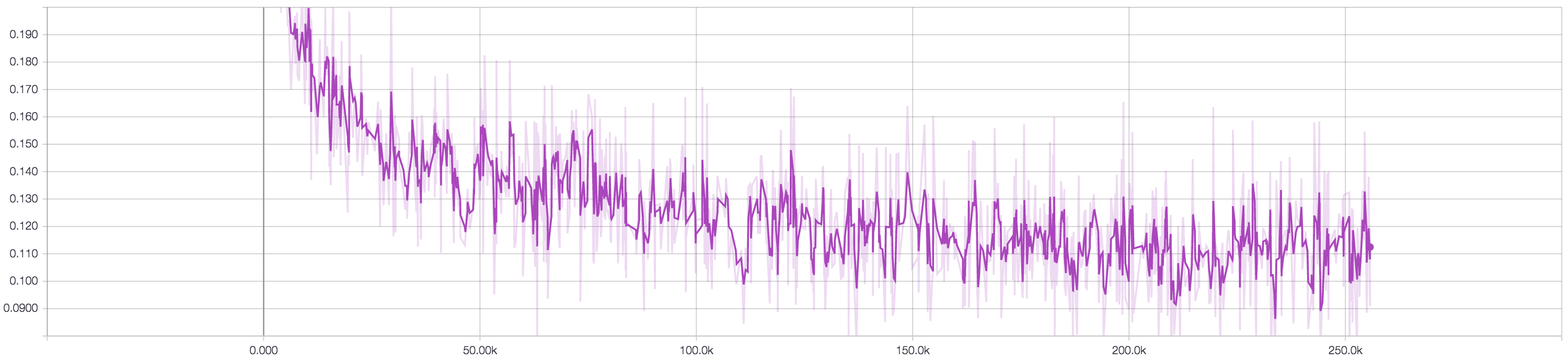}
    \caption{The supervised loss.\\
    source: https://github.com/gitlimlab/SSGAN-Tensorflow}
  \end{minipage}
\end{figure}

\begin{figure}[H]
  \centering
  \begin{minipage}[b]{1\textwidth}
    \includegraphics[scale=0.3, left]{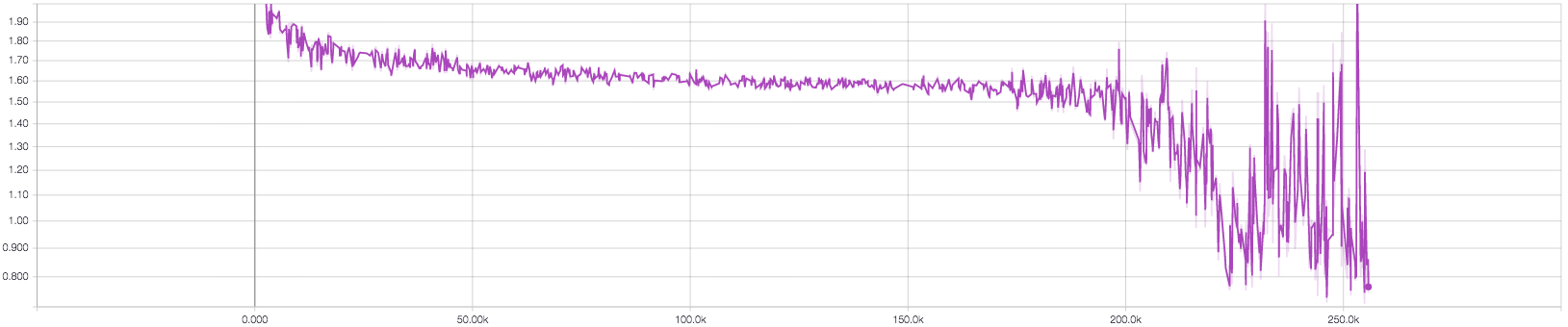}
    \caption{Discriminator total loss.\\
    source: https://github.com/gitlimlab/SSGAN-Tensorflow}
  \end{minipage}
\end{figure}

\begin{figure}[H]
  \centering
  \begin{minipage}[b]{1\textwidth}
    \includegraphics[scale=0.3, left]{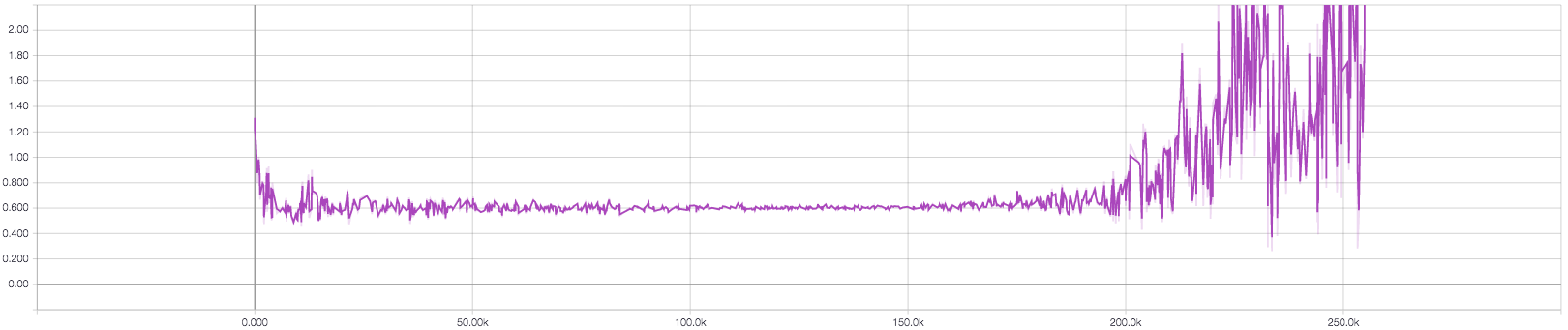}
    \caption{The loss of Generator.\\
    source: https://github.com/gitlimlab/SSGAN-Tensorflow}
  \end{minipage}
\end{figure}

\subsection{Capsule networks}
\forceindent At September of 2017, a new paper \cite{Capsules} was published by Hinton and his team, which introduced a new type of neural network, based on capsules. This new type of neural network could be considered as an improvement of traditional CNNs. The logic behind CNNs is that they can detect and recognize features of an image, but they cannot keep the relative positions between these features, so this make CNNs vulnerable to classify different images as the same. That is what Hinton is trying to solve with Capsule Networks.

\forceindent The use of max pooling after a convolutional layer, leaves behind useful information and ignores relative spatial relationships between features. If the position of an object is shifted by a little in an input image, because of the max pooling, the output of neurons will not change and the network will still detect the object. Capsules solve this problem by saving information about the state of a feature in a vector form.

\forceindent Let's take an example how a face would be detected with capsules, if we assume that there are 2 layers of capsules. In the first layer there are three lower level capsules which detect nose, eyes and mouth. Every capsule encodes some internal states and the probability of existence for each of the above features accordingly. Each of these capsules are then multiplied by some weights, which encode spatial relationship with the higher level capsule, the face. In other words, these output vectors point where the face should be, according to the lower level features(mouth, nose, eyes). If these predictions are close, then it should exist a face. Finally, the final output from the higher level capsule gives the pose and the probability that the face exists.\\

\begin{figure}[H]
  \centering
  \begin{minipage}[b]{1\textwidth}
    \includegraphics[scale=0.4, left]{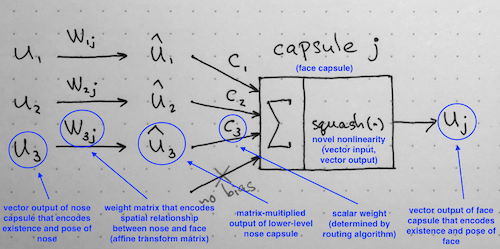}
    \caption{Summary of the internal workings of the capsule.\\
    source: https://medium.com/ai\%C2\%B3-theory-practice-business/understanding-hintons-capsule-networks-part-ii-how-capsules-work-153b6ade9f66 }
  \end{minipage}
\end{figure}

\subsection{CapsuleGAN}
\forceindent Motivated by the better performance of CapsuleNets in compare with CNNs, a new framework which combines GANs and CapsuleNets is proposed in \cite{CapsuleGANS}. The main idea is to substitute the traditional GAN discriminator, which consists of convolutional layers, with capsule layers, with goal to perform the 2-class classification task (real or fake). The architecture of the capsuleGAN discriminator is similar as the one described in 2.8. 

\forceindent The capsuleGAN was implemented and compared with traditional GANs in MNIST and CIFAR-10 databases. There were some images generated by both networks that did not represent any digit(in case of MNIST). However, images generated from capsuleGAN seem to have less diversity in terms of classes and to be cleaner.\\

\begin{figure}[H]
  \centering
  \begin{minipage}[b]{1\textwidth}
    \includegraphics[scale=0.7]{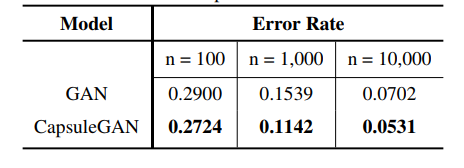}
    \caption{Results of semi supervised classification - MNIST. \\ source :\cite{CapsuleGANS}}
  \end{minipage}
\end{figure}
\begin{figure}[H]
  \centering
  \begin{minipage}[b]{1\textwidth}
    \includegraphics[scale=0.7]{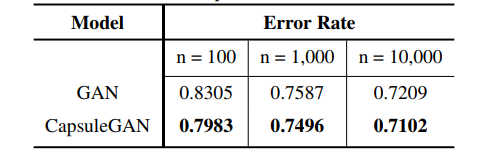}
    \caption{Results of semi supervised classification - CIFAR-10. \\ source :\cite{CapsuleGANS}}
  \end{minipage}
\end{figure}

\section{Architecture of AffWildNet}
\forceindent In paper \cite{kollias15}, there is already some job done in terms of the architecture of a deep learning network that utilizes the existing Aff-Wild database. This includes evaluating architectures based on ResNet L50, VGG Face network and VGG-16 network, an approach where a CNN was used trained only with video frames, and a CNN plus RNN approach which can exploit temporal properties of human behavior\cite{kollias7,kollias14}.

\subsection{VGG}
\forceindent In paper\cite{VeryDeep}, the Visual Geometry Group (VGG) team tries to achieve high accuracy scores by increasing the depth of convolutional networks and more specifically up to 19 layers of weight. The architecture of the proposed networks is similar to the traditional convolution networks, but smaller filters are used in convolutional layers. Their size is usually 3x3, but there is also a case where a 1x1 filter was used and it worked as a linear transformation of the input channels. In general the network is made from a stack of convolution layers, then fully connected layers and a softmax layer.

\forceindent The above architecture was trained and tested in various datasets and had better results than other not so deep networks. This increased accuracy of this type of networks confirms the importance of depth in image classification tasks. \\
\begin{figure}[H]
  \centering
  \begin{minipage}[b]{1\textwidth}
    \includegraphics[scale=1.1]{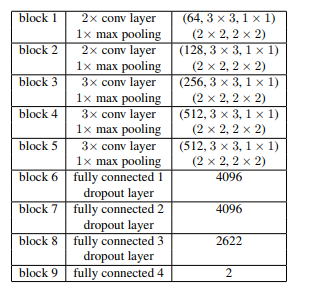}
    \caption{Architecture for CNN network based on VGG-Face/VGG-16. \\ source :\cite{kollias2}}
  \end{minipage}
\end{figure}

\subsection{ResNet}
\forceindent Deep convolutional neural networks have shown great results for image classification in the past years. They have the ability to recognize low/mid/high level features, according to the number of stacked layers existing in the network.  As \cite{ResNet} mentions, the depth of a network is  very crucial, but in very deep networks the problem of vanishing gradient occurs. According to the paper, the proposed solution for this problem is normalized initialization and intermediate normalization layers, which enable networks with many layers to start converging for stochastic gradient descent (SGD) with backpropagation. Another problem arises when deeper networks start converging, which is that accuracy gets saturated and then degrades rapidly. If more layers are added to the network, the training error gets higher instead of lower, so there exists a degradation of the training error.

\forceindent In paper \cite{ResNet}, a solution is addressed to the degradation problem, by introducing a deep residual learning framework.  Instead
of just stacking layers and hope that they directly fit a desired underlying mapping, these layers are explicitly pushed to fit a residual mapping. The desired underlying mapping can be denoted as H(x), so the stacked nonlinear
layers fit another mapping of F(x) := H(x) -- x and the original mapping is recast into F(x)+x. This formulation can be realized by feedforward neural networks with skipping one or more layers, also known as “shortcut connections”. These networks can be still trained by SGD with backpropagation, by using common software libraries.

\forceindent Deep residual nets successfully achieved very challenging results for a number of over 1000 layers. The advantage of these networks is that they are easy to optimize, in contrary with the "plain" networks whose training error gets higher when depth increases. Moreover, the architecture of residual networks can more easily have accuracy gains from greatly increased depth than any other network.\\

\begin{figure}[H]
  \centering
  \begin{minipage}[b]{1\textwidth}
    \includegraphics[scale=1.]{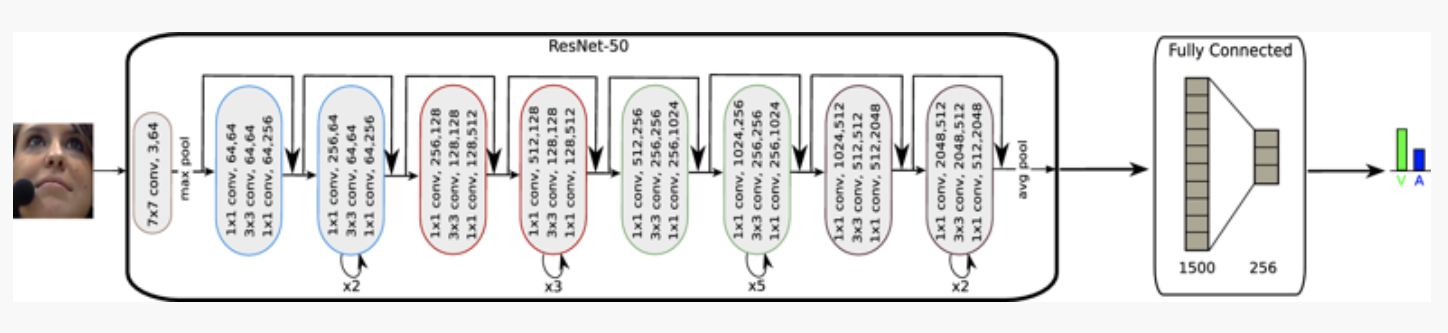}
    \caption{Architecture for CNN network based on ResNet L50. \\ source :\cite{kollias2}}
  \end{minipage}
\end{figure}

\subsection{CNN-RNN}
\forceindent The CNN-RNN architecture was also use in \cite{kollias1,kollias2,kollias3} to evaluate the aff-wild dataset. The input of the RNN network is the output of the first or the second fully connected layer of the respective CNN. Moreover, the RNN consists of one or two hidden layers with 100-150 hidden units. According to \cite{kollias2}, the above network structure provided the best results and parameters optimization was performed.\\

\begin{figure}[H]
  \centering
  \begin{minipage}[b]{1\textwidth}
    \includegraphics[scale=1.2]{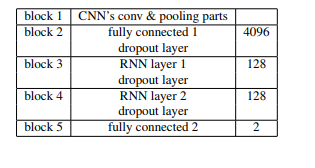}
    \caption{Architecture of CNN-RNN network. \\ source :\cite{kollias2}}
  \end{minipage}
\end{figure}

\subsection{CNN-M}
\forceindent According to \cite{kollias1,kollias2,kollias3}, as there was no previous experiments where it was attempted to predict valence and arousal values in the wild, the following method was followed, based on the CNN-M \cite{CNN_M} network. More specifically, the pre-trained CNN-M network with the FaceValue \cite{fv} dataset was used as starting structure and transfer learning was performed, with the exact structure shown in Figure 2.23. The fine-tuning could be performed with two possible ways. First, by freezing the CNN part of the network and performing fine-tuning of the fully connected layers. Second, by performing fine-tuning to the whole network.
\begin{figure}[H]
  \centering
  \begin{minipage}[b]{1\textwidth}
    \includegraphics[scale=1.3]{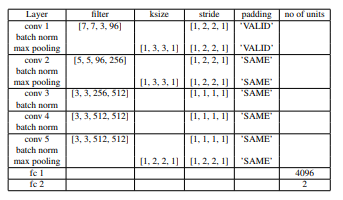}
    \caption{The architecture of the CNN-M model. \\ source :\cite{kollias1,kollias2,kollias3}}
  \end{minipage}
\end{figure}

\subsection{Results Of The Above Architectures}
\begin{figure}[H]
  \centering
  \begin{minipage}[b]{1\textwidth}
    \includegraphics[scale=1.]{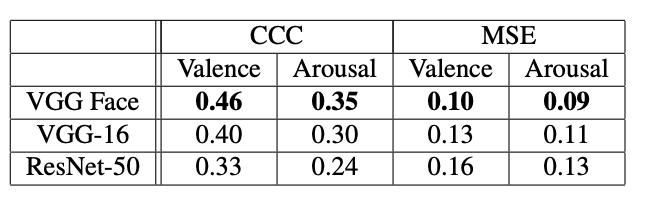}
    \caption{Results of VGGFace, VGG-16 and ResNet50 architectures, that were described above. \\ source :\cite{kollias2}}
  \end{minipage}
\end{figure}
\begin{figure}[H]
  \centering
  \begin{minipage}[b]{1\textwidth}
    \includegraphics[scale=1.]{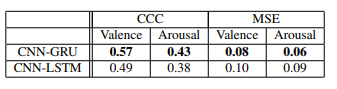}
    \caption{Result of the CNN-RNN architecture that described above, with GRU and LSTM cells. \\ source :\cite{kollias2}}
  \end{minipage}
\end{figure}
\begin{figure}[H]
  \centering
  \begin{minipage}[b]{1\textwidth}
    \includegraphics[scale=1.]{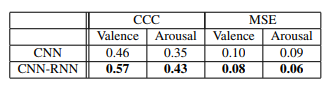}
    \caption{Best results of the above CNN and CNN-RNN architectures. \\ source :\cite{kollias2}}
  \end{minipage}
\end{figure}
\begin{figure}[H]
  \centering
  \begin{minipage}[b]{1\textwidth}
    \includegraphics[scale=1.]{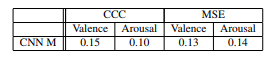}
    \caption{The results of the modified CNN-M model. \\ source :\cite{kollias1,kollias2,kollias3}}
  \end{minipage}
\end{figure}
\forceindent All the figures of this section(2.24, 2.25, 2.26, 2.27) are very important for this project, because they are going to be example results that will be compared with the results of the models of this project. The CCC value will be used for this project, so this value is the one that is more important in these figures. It can be observed that for the RNN and CNN architectures the best mean \textbf{CCC} is approximately \textbf{0.5} and \textbf{0.4} respectively. For the modified CNN-M model the mean CCC is approximately \textbf{0.13}.
\chapter{Dataset}
\section{Kind of data needed}
\forceindent For the present project, data from video are needed and more specifically videos that are not made for any specific task of emotions recognition, in order to satisfy the "in the wild" concept. Hence, like in \cite{kollias1,kollias2,kollias3}, the videos are gathered from YouTube and they are usually videos which can be found with the keyword "Reaction". Moreover, they contain people who usually react in unexpected plots, exciting situations, etc. 

\forceindent These videos will be the input to various Deep Learning models, in the format of images (extracted frames for the videos), and the output will be valence and arousal values. Hence, it is important that our videos have the best possible quality, because it had be shown that quality affects the performance of Neural Networks \cite{quality}. So, we expect the size of the data to be large for some videos/frames, but as long as the quality is high, data size is not a concern.
\\
\begin{figure}[H]
  \centering
  \begin{minipage}[b]{1\textwidth}
    \includegraphics[scale=0.78, left]{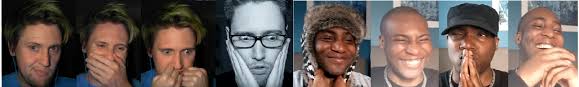}
    \caption{Some frames from the aff-wild database.\\
    source: \cite{kollias1,kollias2,kollias3}}
  \end{minipage}
\end{figure}
\section{Data gathering}

\forceindent With the \cite{ytdown} library we were able to download videos in best quality available, by running a specific command in command line for every video. Because of the big number of videos, a python script was created, which reads the videos link from a text file, ran the needed command automatically for every video and finally saved the video file in a specific directory. The total size of all gathered videos was 5 GB. 
\section{Data pre-processing}

\forceindent Before the annotation procedure, the downloaded videos have to be preprocessed for two reasons. First, they need to all have the same frame rate. Second, they need to include only scenes where people react and not irrelevant scenes(e.g. an advertisement).

\forceindent In order to specify the frame rate which all the videos need to have, the following procedure was implemented. The frame rate of every video was found and an histogram with the frame rate in FPS was created, in order to observe the distribution of the frame rate in our videos. A python script was used for this scope with the ffprobe\cite{ffprobe} command, which gathers information of multimedia streams(e.g. a video), including the FPS.

\begin{figure}[H]
  \centering
  \begin{minipage}[b]{1\textwidth}
    \includegraphics[scale=0.8, left]{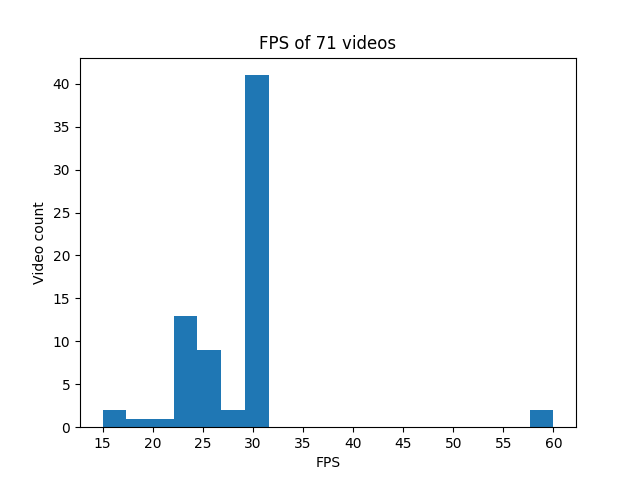}
    \caption{The FPS value for every video.}
  \end{minipage}
\end{figure}
\capstartfalse
\begin{table}[H]
\centering
\begin{tabular}{|c|c|c|}
\hline
                  &30 FPS & No 30 FPS\\
                  \hline
Number of videos  & 41  & 30 \\
\hline
Duration          & 3:48:00  & 2:13:00\\
\hline
\end{tabular}
\end{table}
\capstarttrue
\begin{table}[H]
  \caption{Number of videos and duration for videos that are 30 fps and for videos that are not 30 fps.}
\end{table}
\forceindent Most videos had near or exactly 30 FPS, so the value of 30 was decided to be the FPS value for all the videos of the database. \textbf{At this point, 71 videos were gathered} and because some of them contain 2 people or more, they will be separated in a next stage. So, \textbf{the number of videos is expected to increase.}

\forceindent For the second part of data pre-processing, the video trimming part, OpenShot was used \cite{OS}. The main purpose was to delete scenes that are not useful for the project. These scenes could be an ad, an intro or an outro. In this case, every video was processed individually. \cite{OS} has the functionality to export the videos in a selected frame rate for every selected quality. So, instead of creating another procedure for the frame rate conversion, the new fps value was selected through the \cite{OS}.

\forceindent So, the dataset that occurred from the above steps includes videos with the same frame rate and with frames that include only human faces. Some videos had to be separated, because they included individual scenes with different people. For example, video 45 that shows the reaction of 7 people inside a car in different scenes for every person, it was separated in 7 videos. So, \textbf{78 videos} are ready to be fed to the annotation procedure.
\section{Data annotation}

\forceindent For data annotation the same tool as \cite{kollias2} was used. Every video was annotated in terms of valence and arousal individually. With the tool and the joystick provided and with some examples of annotated frames, a value for each valence and arousal was given for a constant time-step. The output of the annotation procedure is two text files for every video with these 2 series of values.  
\\
\begin{figure}[H]
  \centering
  \begin{minipage}[b]{1\textwidth}
    \includegraphics[scale=1.5, left]{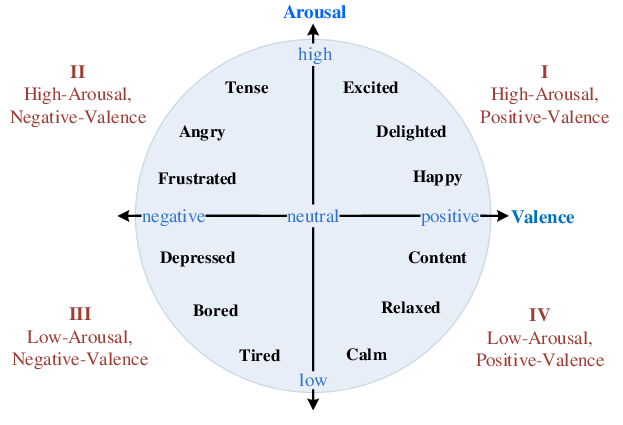}
    \caption{Guidance image for annotation.\\
    image source: https://www.researchgate.net/figure/Two-dimensional-valence-arousal-space\_fig1\_304124018\\}
  \end{minipage}
\end{figure}

\forceindent The value of valence or arousal in a video can rapidly change. For example, if a person in calm state sees something sad, he/she will develop negative valence with low arousal. Since the annotation happens in real time, the annotator must be able to fast adjust the values of valence or arousal according to what he/she sees. Else, the annotation values would have a delay and they would be not assigned to the right frame. The joystick contributes a lot to solving this issue, because it is easy to move and it has scales of motion(small move for small motion - big move for big motion). Therefore, the only delay between the frame and the corresponding annotation value is the reaction time of the annotator.

\begin{figure}[H]
  \centering
  \begin{minipage}[b]{0.45\textwidth}
    \includegraphics[scale=0.45, left]{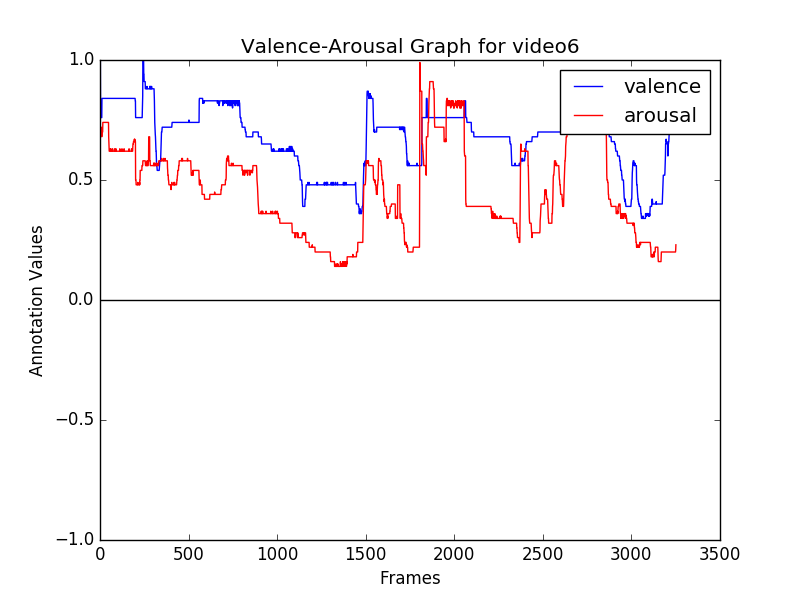}
    \caption{Valence-Arousal annotations for video6.}
  \end{minipage}
  \hfill
  \begin{minipage}[b]{0.45\textwidth}
    \includegraphics[scale=0.45, left]{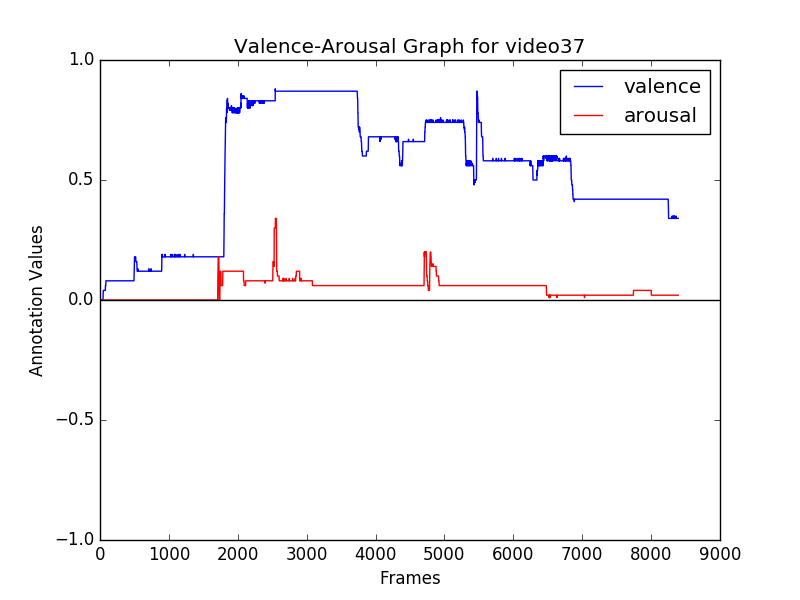}
    \caption{Valence-Arousal annotations for video37.}
  \end{minipage}
\end{figure}

\begin{figure}[H]
  \centering
  \begin{minipage}[b]{0.45\textwidth}
    \includegraphics[scale=0.45, left]{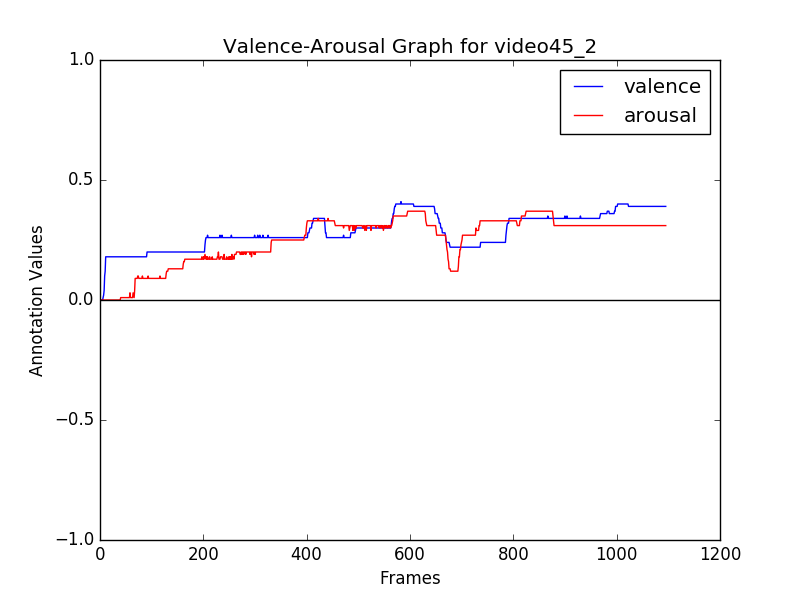}
    \caption{Valence-Arousal annotations for video45\_2.}
  \end{minipage}
  \hfill
  \begin{minipage}[b]{0.45\textwidth}
    \includegraphics[scale=0.45, left]{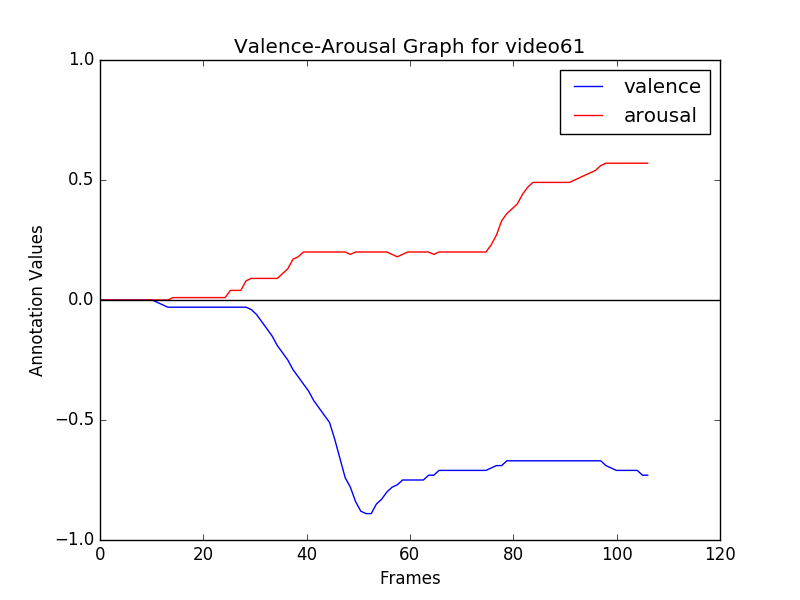}
    \caption{Valence-Arousal annotations for video61.}
  \end{minipage}
\end{figure}

\forceindent The choice of the videos was not random. A Neural Network \cite{kollia2009interweaving} needs a good distribution of labels in order to be trained properly. So, videos with a broad range of values for valence and arousal had to be chosen. This means that all the emotions are wanted to be exposed to this experiment. Since the source for the data is the web, the most common values for valence and arousal were those near zero, which mostly correspond to a person who is calmly talking. Nevertheless, as it can be seen in the following figure, the distribution was not as good as expected. That is because most of the valence and arousal values of the frames are gathered near zero.

\begin{figure}[H]
  \centering
  \begin{minipage}[b]{1\textwidth}
    \includegraphics[scale=0.7, left]{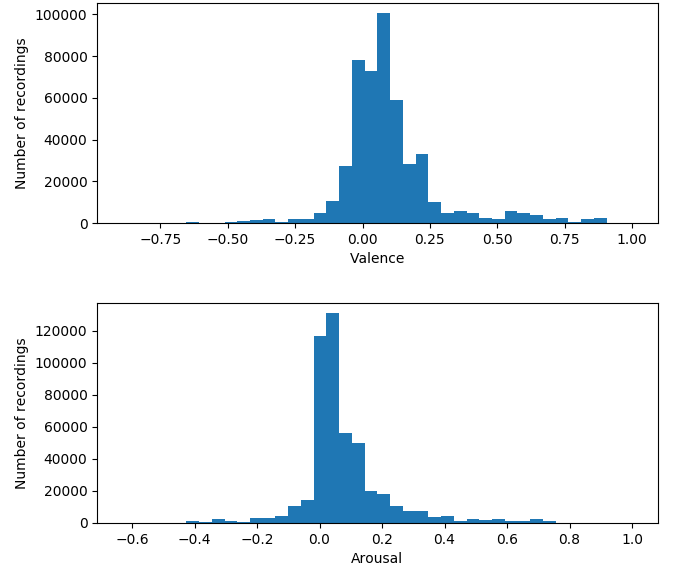}
    \caption{Valence and arousal distribution for the whole dataset (479.022 frames before face detection).}
  \end{minipage}
\end{figure}

\forceindent It should be mentioned that videos that contain 2 or more people in the same scene, are labeled for each person. For example video55 contains one man and one woman and their corresponding annotation files are 2 and they are named video55\_left.txt and video55.txt, where the first file is the person on left as we look at the screen. In this case the video counts as two videos.
\section{Extra data}
\forceindent Since the current distribution of valence and arousal values is not the desired one, more data will be added to the dataset. For the new videos the same procedure as before will be followed. The goal for the new dataset is to have a broad range of annotation values, so when it get merged with the initial database, to have a good overall distribution.

\forceindent The first thing to do after downloading \textbf{the new 21 videos} is to check the FPS distribution, in order to confirm that 30 fps is still the appropriate value to keep. 
\begin{figure}[H]
  \centering
  \begin{minipage}[b]{1\textwidth}
    \includegraphics[scale=0.8, left]{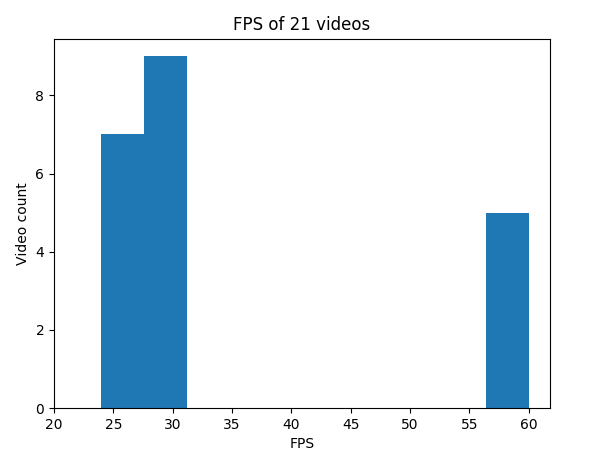}
    \caption{The FPS value for every video (for the new videos).}
  \end{minipage}
\end{figure}
\capstartfalse
\begin{table}[H]
\centering
\begin{tabular}{|c|c|c|}
\hline
                  &30 FPS & No 30 FPS\\
                  \hline
Number of videos  & 9  & 12 \\
\hline
Duration          & 0:52:00  & 1:49:00\\
\hline
\end{tabular}
\end{table}
\capstarttrue
\begin{table}[H]
  \caption{Number of videos and duration for videos that are 30 fps and for videos that are not 30 fps(for the new videos).}
\end{table}
\forceindent Still more than the half data are 30 fps and there are many videos near 30 fps, so that is the value that will be kept. After processing the videos with the same tool as before, 2 new videos came up, so \textbf{in total 23 new videos} were added to the initial database.

\forceindent Last thing that it needs to be checked is if the annotation distribution is the desired one. According to the figure below, clearly the values are better distributed in low and high valence and arousal values.

\begin{figure}[H]
  \centering
  \begin{minipage}[b]{1\textwidth}
    \includegraphics[scale=0.7, left]{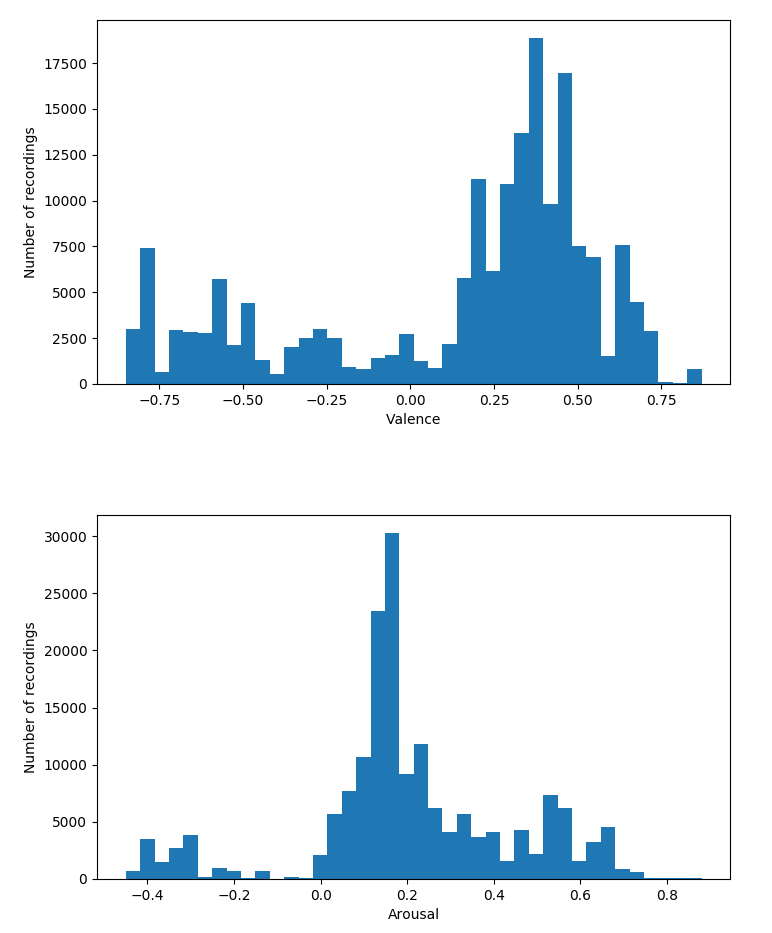}
    \caption{Valence and arousal distribution for the new dataset (85.556 frames before face detection).}
  \end{minipage}
\end{figure}

\section{Face detection}
\forceindent The last step before the dataset is ready to be fed to any neural network is to detect and crop the faces that exist on every frame for every video. So, the task is to detect face in an image/frame. There are several ways to achieve face detection, but for this case the menpo library \cite{menpo} was used. More specifically, the menpodetect pack was used with the ffld2 detector. The ffld2 detector achieves frontal face detection using the DPM Baseline model provided by Mathias et. al. Faces are being detected even when hands are covering a part of them or if the person wear glasses. That is very desirable in this project, in order to keep the "in the wild" concept.\\

\begin{figure}[H]
  \centering
  \begin{minipage}[b]{0.4\textwidth}
    \includegraphics[scale=0.2, left]{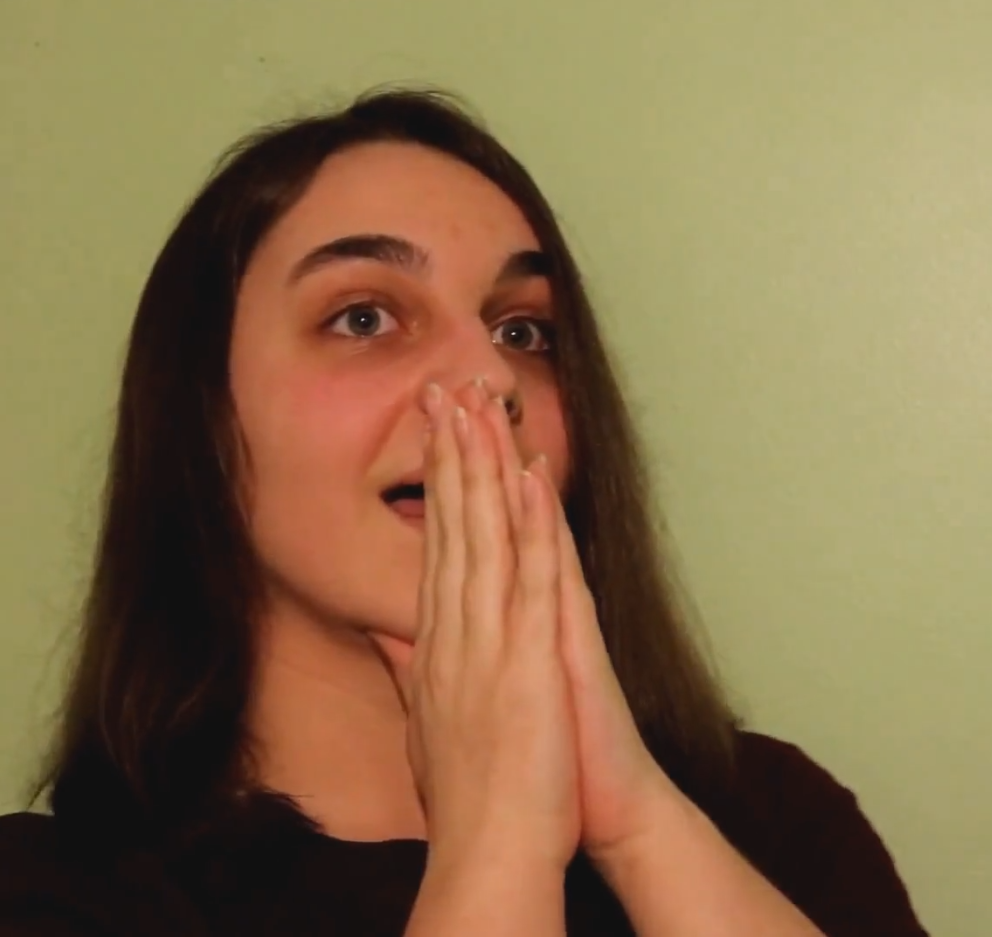}
    \caption{Original frame.}
  \end{minipage}
  \hfill
  \begin{minipage}[b]{0.5\textwidth}
    \includegraphics[scale=0.3, left]{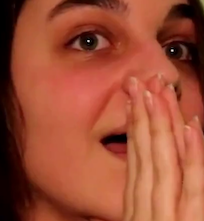}
    \caption{Frame after face detection.}
  \end{minipage}
\end{figure}

\forceindent Another thing that was important about this part of the data processing, was to keep the loss of frame due the face detection procedure low. The frames of the original videos before the frames detection were 564.578. After the face detection the resulting dataset consisted of 507.208 frames, meaning that there was a 10\% loss, which is good enough to proceed to the next steps.

\begin{figure}[H]
  \centering
  \begin{minipage}[b]{1\textwidth}
    \includegraphics[scale=0.43, left]{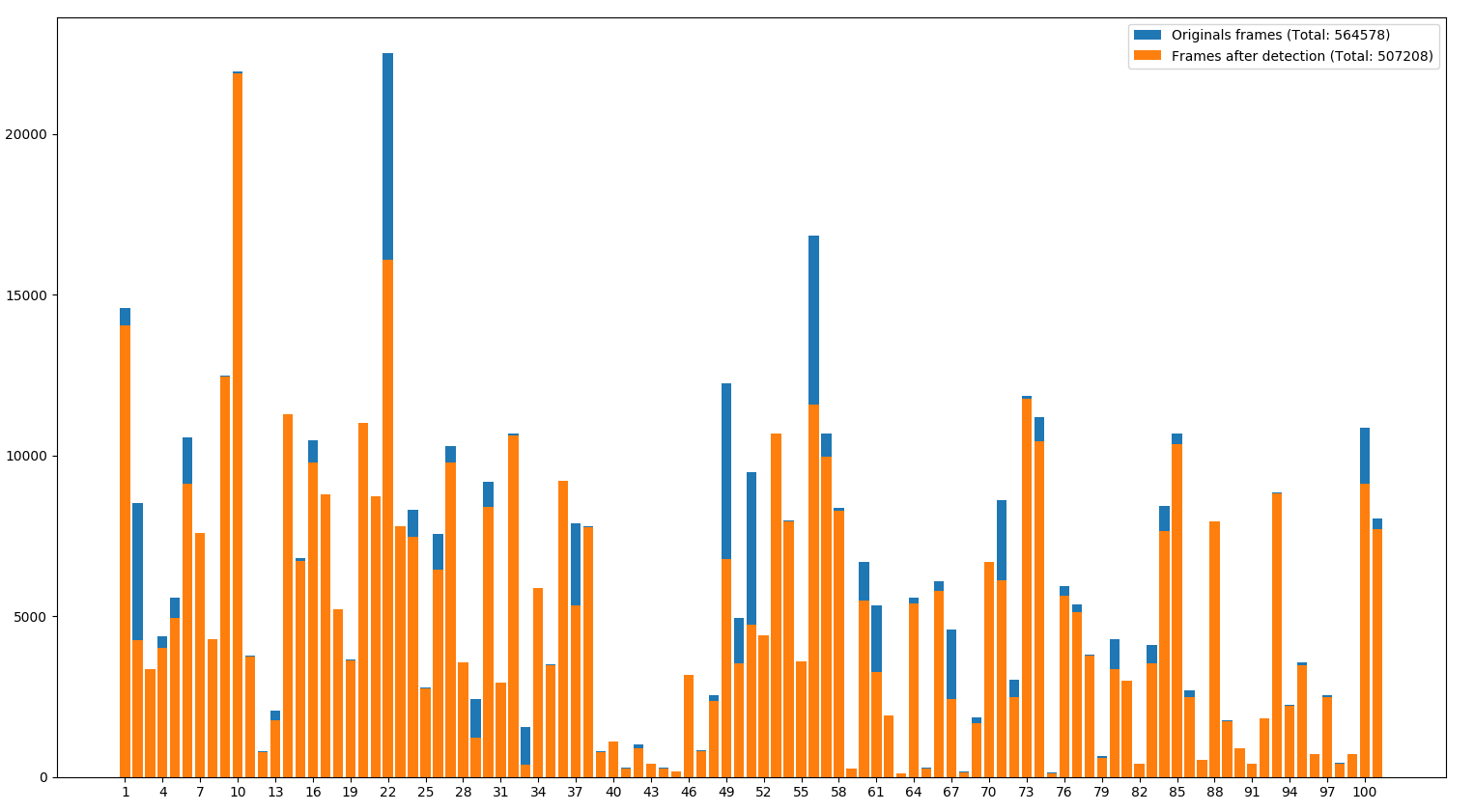}
    \caption{Frames before and after the face detection for every video. (Videos which include 2 people have been counted in the same bar twice.)}
  \end{minipage}
\end{figure}

\forceindent In order to have a consistent way of saving the frames which the detector gave as output, the following way of saving was followed. Every video has its own directory named with the video name(e.g. video1, video2, etc) and it contains the detected frames with the number of frame as name in 6-digit format (e.g. 0000001.png, 0000021.png, etc).
\section{Annotations to labels conversion}
\forceindent The output of the annotation tool was in the following format: \\

\capstartfalse
\begin{table}[H]
\centering
\begin{tabular}{lllllllllll}
\multicolumn{2}{c}{Valence} & \multicolumn{7}{l}{\multirow{7}{*}{}} & \multicolumn{2}{c}{Arousal} \\
Seconds       & Values      & \multicolumn{7}{l}{}                  & Seconds       & Values      \\
0.010         & 0           & \multicolumn{7}{l}{}                  & 0.016         & 0           \\
0.541         & -408        & \multicolumn{7}{l}{}                  & 0.037         & 0           \\
0.556         & -432        & \multicolumn{7}{l}{}                  & 0.116         & 0           \\
0.576         & -448        & \multicolumn{7}{l}{}                  & 0.176         & -37         \\
0.587         & -448        & \multicolumn{7}{l}{}                  & 0.218         & -116       
\end{tabular}
\end{table}
\capstarttrue
\begin{table}[H]
  \caption{Output of annotation tool for valence and arousal for video89}
\end{table}

\forceindent There are several problems that have to be solved, in order to convert the format in Table 3.1 into format of labels that can be fed to Neural networks.

\forceindent First, the time step that the software captures values for valence and arousal is different and it is captured in seconds. This is a problem, because our dataset contains frames of videos (30 frames for every second), so a way must be found to match these two values. Linear interpolation \cite{interp} for valence and arousal sequences individually was used for this task. More specifically, since there is a standard frame rate for every video it was easy to create the appropriate sequences to feed into python interp(x, xp, fp) function, according to the documentation \cite{interp_code}:\\

x: The x-coordinates at which to evaluate the interpolated values. A label for every frame is wanted. If total frames of video are set as 'frames':\\
\[ 1, 2, 3 ,4, ..., frames - 1, frames\]

xp: The x coordinates of data points. Since the previous sequence is defined in frames, this should also be defined in frames. So it will be defined as the values in seconds from annotation tool multiplied by 30 (fps):\\
(in case of valence of video 89)
\[0.010*30, 0.541*30, 0.556*30, 0.576*30,...,74.754*30\]

fp: The y-coordinates of the data points. Here, just the values of the sequence for valence or arousal will be entered:\\
(in case of valence of video 89)
\[0,-408,-432,-448,...,-828\]

\forceindent Now that there is a valance-arousal value for every frame, these values should be scaled. The current range of the values is [-1000,1000] and the desired range is [-1,1], so all values will be divided by 1000.

\forceindent The last step for creating the labels, is to create a text file that contains the valence-arousal values, which correspond to every frame that has been detected with a face. As mentioned before, every video is saved in a directory with the video's name. So, for every directory the frame number are found and the corresponding valence-arousal values are printed in the text file. Hence, every video has its own labels file named with the video's name (e.g. video89.txt).
\section{Final dataset}
\forceindent The final dataset consists of 507.208 frames gathered from 106 videos and its size is approximately 48GB. These frames includes approximately 150 faces of different people.

\forceindent The resulting frames can be considered as very clean data for any "in the wild" face related machine learning task. The annotation was created by one person, so it is not as objective as in the "Aff-wild" database. Nevertheless, the dataset has the potential to work right with the experiments that are going to take place.
\capstartfalse
\begin{table}[H]
\centering
\begin{tabular}{|c|c|c|c|c|}
\hline
                  &Videos & Frames & no of males & no of females\\
                  \hline
Stage 1          & 71  & 479.022 & 53& 30\\
Stage 2          & 78  & 479.022 & &\\
\hline
Stage 3          & 21  & 85.556 & 15& 8\\
Stage 4          & 23  & 85.556 & &\\
\hline
Stage 5          & 101  & 564.578 & 68  & 38\\
Stage 6          & 106  & 507.208 &  & \\
\hline
\end{tabular}
\end{table}
\capstarttrue
\begin{table}[H]
  \caption{Number of videos,frames and number of males and females for every stage.\\
  Stage 1: Initial database before processing \\
  Stage 2: Initial database after processing (same fps, just separated videos)\\
  Stage 3: Extra database before processing\\
  Stage 4: Extra database after processing (same fps, just separated videos)\\
  Stage 5: Total database before face detection\\
  Stage 6: Final database after face detection\\}
\end{table}
\begin{figure}[H]
  \centering
  \begin{minipage}[b]{0.19\textwidth}
    \includegraphics[height=3.8cm, width=3.2cm, left]{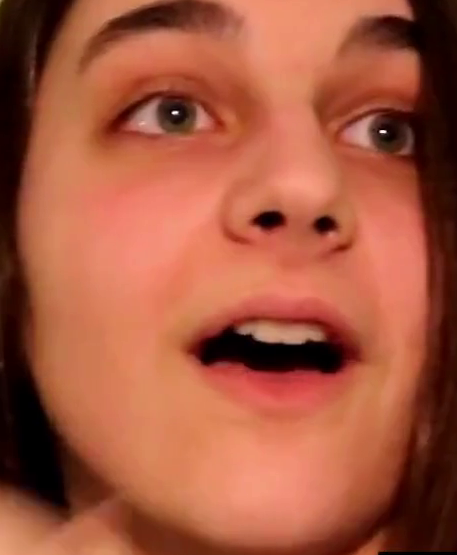}
  \end{minipage}
  \hfill
  \begin{minipage}[b]{0.19\textwidth}
    \includegraphics[height=3.8cm, width=3.2cm, left]{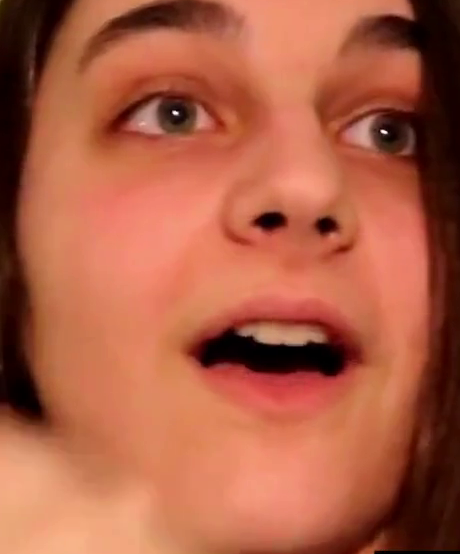}
  \end{minipage}
  \hfill
  \begin{minipage}[b]{0.19\textwidth}
    \includegraphics[height=3.8cm, width=3.2cm, left]{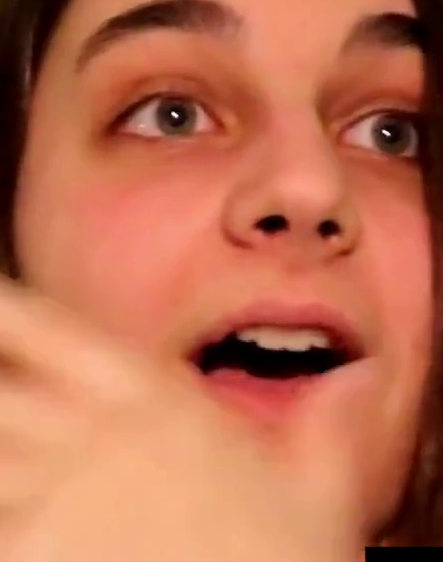}
  \end{minipage}
  \hfill
  \begin{minipage}[b]{0.19\textwidth}
    \includegraphics[height=3.8cm, width=3.2cm, left]{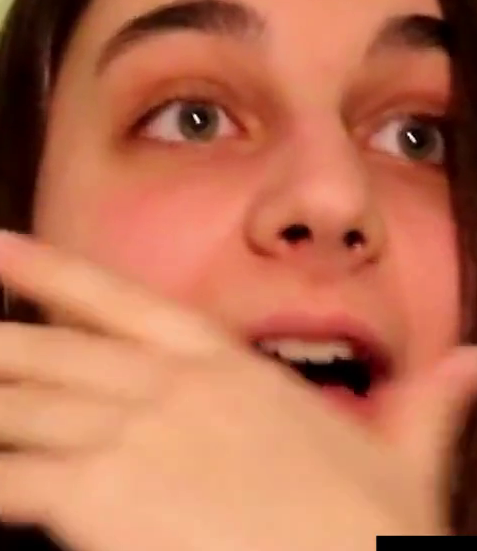}
  \end{minipage}
  \hfill
  \begin{minipage}[b]{0.19\textwidth}
    \includegraphics[height=3.8cm, width=3.2cm, left]{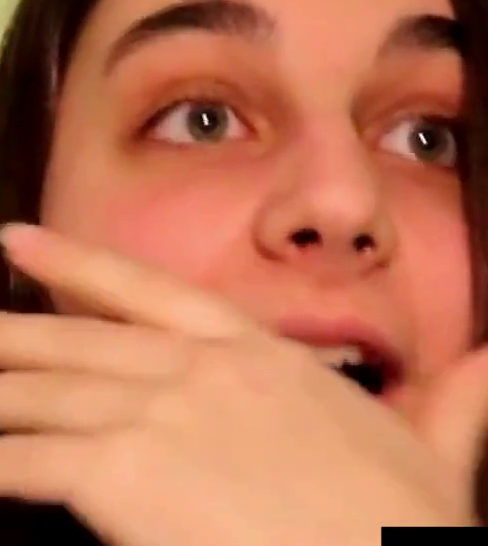}
  \end{minipage}
\end{figure}
\begin{figure}[H]
  \centering
  \begin{minipage}[b]{0.19\textwidth}
    \includegraphics[height=3.8cm, width=3.2cm, left]{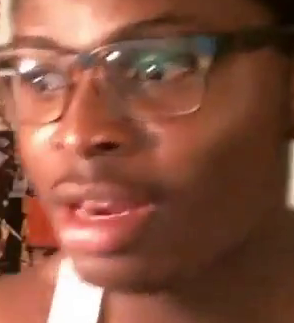}
  \end{minipage}
  \hfill
  \begin{minipage}[b]{0.19\textwidth}
    \includegraphics[height=3.8cm, width=3.2cm, left]{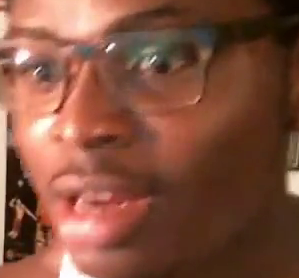}
  \end{minipage}
  \hfill
  \begin{minipage}[b]{0.19\textwidth}
    \includegraphics[height=3.8cm, width=3.2cm, left]{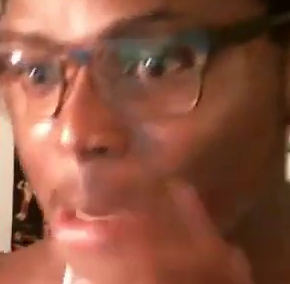}
  \end{minipage}
  \hfill
  \begin{minipage}[b]{0.19\textwidth}
    \includegraphics[height=3.8cm, width=3.2cm, left]{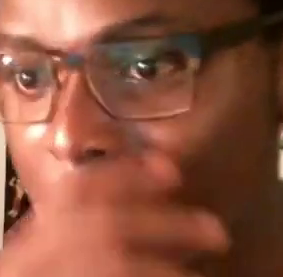}
  \end{minipage}
  \hfill
  \begin{minipage}[b]{0.19\textwidth}
    \includegraphics[height=3.8cm, width=3.2cm, left]{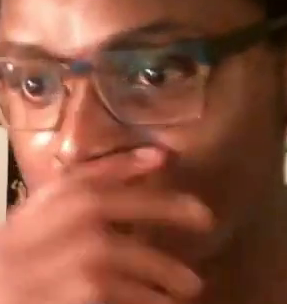}
  \end{minipage}
\end{figure}
\begin{figure}[H]
  \centering
  \begin{minipage}[b]{0.19\textwidth}
    \includegraphics[height=3.8cm, width=3.2cm, left]{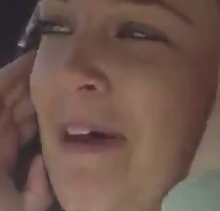}
  \end{minipage}
  \hfill
  \begin{minipage}[b]{0.19\textwidth}
    \includegraphics[height=3.8cm, width=3.2cm, left]{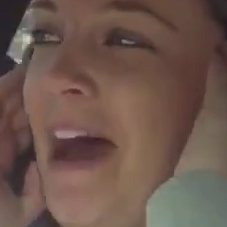}
  \end{minipage}
  \hfill
  \begin{minipage}[b]{0.19\textwidth}
    \includegraphics[height=3.8cm, width=3.2cm, left]{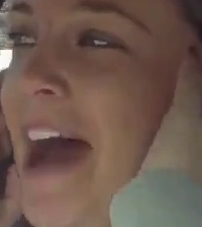}
  \end{minipage}
  \hfill
  \begin{minipage}[b]{0.19\textwidth}
    \includegraphics[height=3.8cm, width=3.2cm, left]{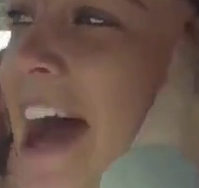}
  \end{minipage}
  \hfill
  \begin{minipage}[b]{0.19\textwidth}
    \includegraphics[height=3.8cm, width=3.2cm, left]{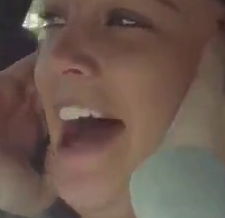}
  \end{minipage}
\end{figure}
\begin{figure}[H]
  \centering
  \begin{minipage}[b]{0.19\textwidth}
    \includegraphics[height=3.8cm, width=3.2cm, left]{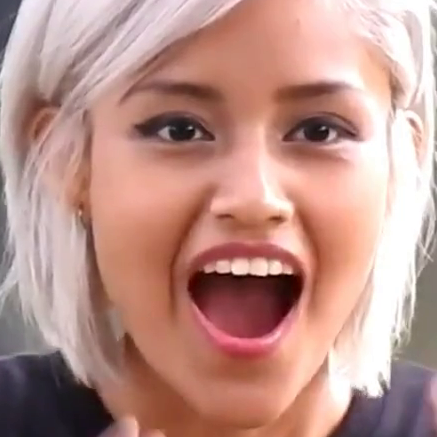}
  \end{minipage}
  \hfill
  \begin{minipage}[b]{0.19\textwidth}
    \includegraphics[height=3.8cm, width=3.2cm, left]{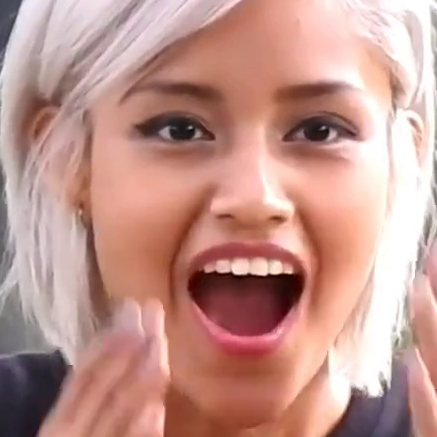}
  \end{minipage}
  \hfill
  \begin{minipage}[b]{0.19\textwidth}
    \includegraphics[height=3.8cm, width=3.2cm, left]{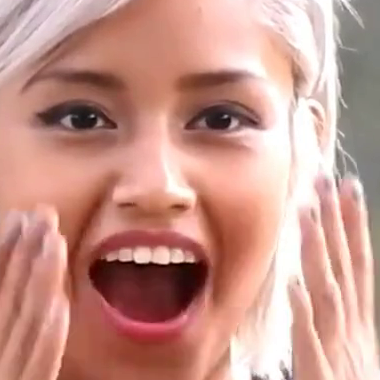}
  \end{minipage}
  \hfill
  \begin{minipage}[b]{0.19\textwidth}
    \includegraphics[height=3.8cm, width=3.2cm, left]{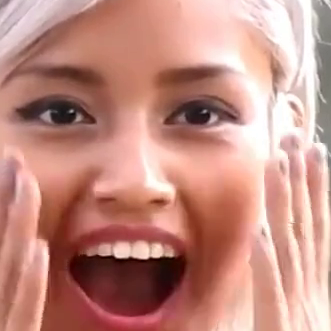}
  \end{minipage}
  \hfill
  \begin{minipage}[b]{0.19\textwidth}
    \includegraphics[height=3.8cm, width=3.2cm, left]{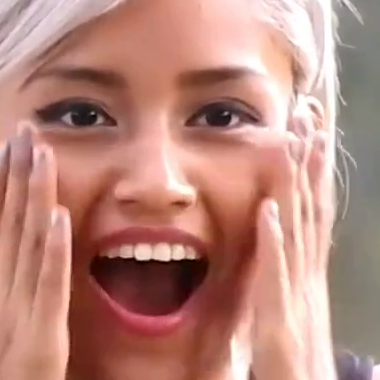}
  \end{minipage}
  \caption{Some challenging sequences of frames.}
\end{figure}
\chapter{Neural Network Models}
\section{Loss function for regression}
\forceindent The dataset that was created in the previous steps is ready to be fed in Neural Networks. Since there is no evidence that the particular dataset will provide accurate predictions for valence and arousal, it has to be first trained in a small network that has been used before. 

\forceindent No matter if a CNN or a GAN is being trained, the loss of the network will include a supervised loss, which will present how accurate the predictions are in terms of valence and arousal. The loss function that will be used for this task is the Concordance Coefficient Correlation (CCC). CCC is defined as:\\
\[ccc=\dfrac{2s_{xy}}{s_x^2+s_y^2 +(\bar{x}-\bar{y})^2}\] \\
where $s_x^2$ and $s_y^2$ are the variances of the predicted and ground truth values respectively, $\bar{x}$ and $\bar{y}$ are the corresponding mean values and $s_{xy}$ is the respective covariance value. In case that predictions and ground truth values are identical CCC gets the value of 1 and in case that these values are completely uncorrelated the values gets the value of 0. Optimizing the loss needs to be a minimization problem, so the loss that will be used in the Neural Networks is 1 - CCC.
\section{Data Split}
\subsection{Testing Set}
\forceindent In order to create a testing set, some videos were subtracted from the initial dataset and they were defined as testing set. Those videos were video7, video19, video25, video45\_1, video45\_2, video45\_3, video45\_4, video45\_5, video45\_6, video45\_7, video48, video62, video72, video73 and they include in \textbf{total 24.226 frames}, with 8 males and 6 females and their valence-arousal distribution is shown in the next figure.
\begin{figure}[H]
  \centering
  \begin{minipage}[b]{1\textwidth}
    \includegraphics[scale=0.6]{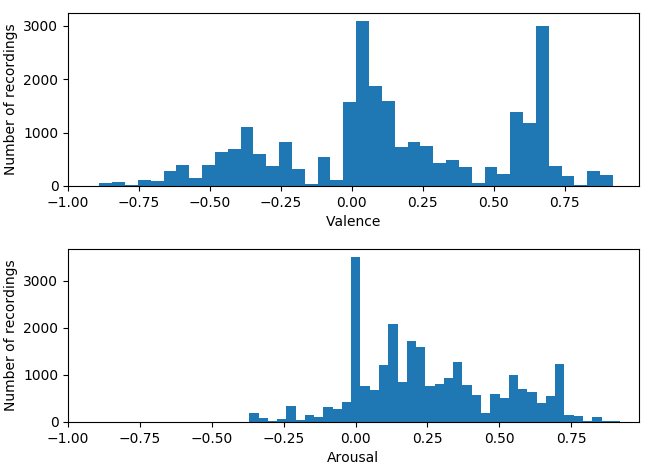}
    \caption{The valence-arousal distribution of the testing set.}
  \end{minipage}
\end{figure}
\subsection{Training With Small Dataset}
\forceindent The extra dataset that was created is defined as small dataset. It includes 80.280 frames, with 15 males and 8 females and its distribution is shown in figure 3.10.
\subsection{Training With Big Dataset}
\forceindent The big dataset does not include the frames from the testing dataset, while it includes the frames from the initial dataset plus the frames from the small dataset. So, there are in total 482.982 frames in the dataset, with 60 males and 32 females and the valence-arousal distribution is shown in the next figure.
\begin{figure}[H]
  \centering
  \begin{minipage}[b]{1\textwidth}
    \includegraphics[scale=0.6]{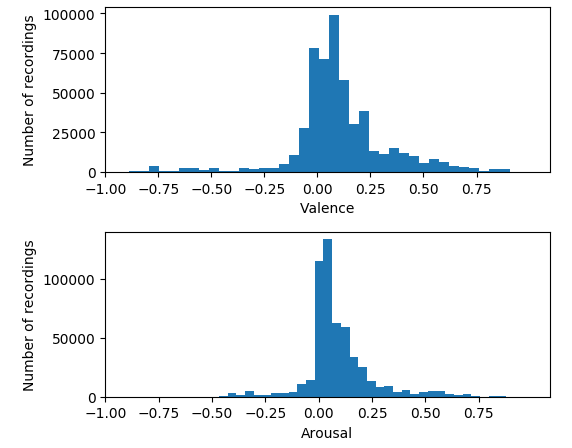}
    \caption{The valence-arousal distribution of the big dataset.}
  \end{minipage}
\end{figure}
\section{Testing The Database In A CNN}
\forceindent In order to ensure that the dataset creates in the previous steps is well made, it will be tested in an already tried CNN architecture. A simple architecture as CNN\_M was chosen for this task and only 110.000 (80.000 for training and 30.000 for testing) frames were used, in order to achieve overfitting and make sure that the labeling makes sense.
\forceindent The network architecture that was used for this scope, was the modified CNN-M \cite{CNN_M} network from paper \cite{kollias1,kollias2,kollias3}. \\
\begin{figure}[H]
  \centering
  \begin{minipage}[b]{1\textwidth}
    \includegraphics[scale=1.3]{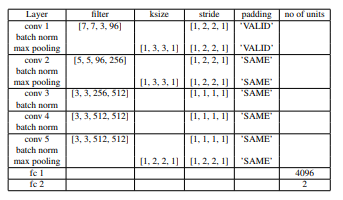}
    \caption{The architecture of the model.}
  \end{minipage}
\end{figure}

\forceindent The small dataset is not derived randomly from the big one. Even with less images, a good valence-arousal distribution is needed for the training dataset, in order to evaluate the model. This distribution is shown in figure 3.10.

\section{Regression GAN}
\subsection{Format of Images}
\forceindent The first thing that had to be done for the gathered data was to set a constant image size, which the images had to be reshaped before they were fed to the model. Since the database was big and that could make the training procedure really slow in case of a big image size, the ideal size was decided by finding the minimum size of all the images in the dataset. The minimum size was 69x69, so a size of 64x64, which is already used in many famous GANs implementations, was chosen and all the images were reshaped to this size.
 
\begin{figure}[H]
  \centering
  \begin{minipage}[b]{1\textwidth}
    \includegraphics[scale=0.41]{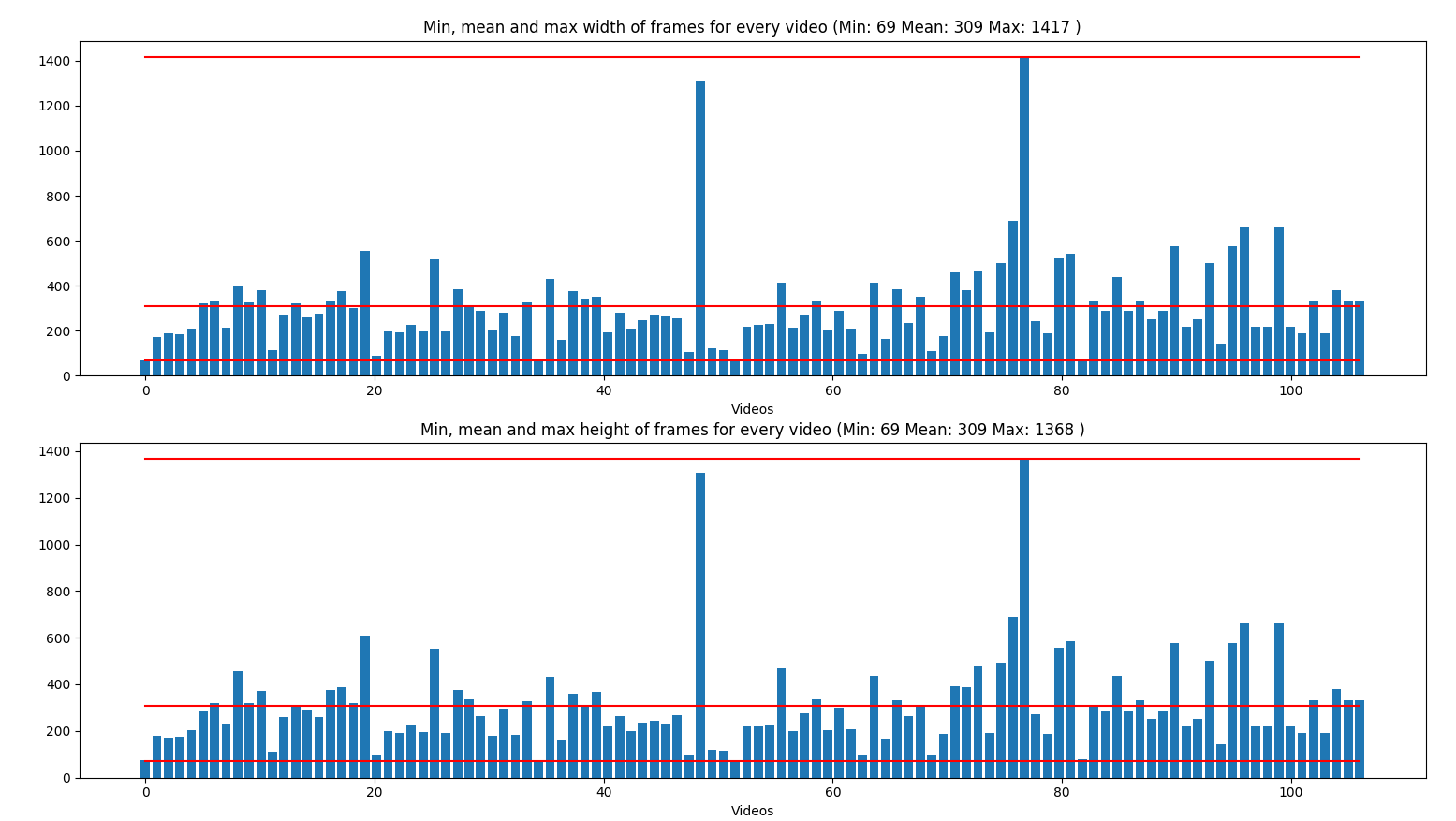}
	\caption{Minimum, mean and maximum width and height for every video (The mean for all the frames of a video was used)}
  \end{minipage}
\end{figure}
\forceindent No data augmentation happened nor noise was added to the data. The only change was, that all the images which had an initial range of values [0, 255], were scaled to [-1, 1]. That was proposed in DCGAN paper \cite{DCGAN}. In the next section, where the structure of the generator is presented, it will be stated that the activation function after the last convolution of the generator is the Tanh function, which outputs values in range [-1, 1]. Since both the real images and the generated samples have to be fed to the discriminator, they have to be in the same range of values, which is [-1, 1].
\subsection{Generator}
\forceindent For the generator, the structure of DCGAN generator was followed. As input, a noise generated from uniform distribution with range from -1 to 1 and dimension 100 was used. In order to connect the input with the convolution layers(project and reshape), a fully connected layer was used. Since the desired starting size for the input of convolution layers was [batch\_size, 4, 4, 512], the output dimensions of the fully connected layer had to be $4*4*512$.

\forceindent After projecting and reshaping the input, four strided convolution layers followed with kernel size [5, 5] and stride [2, 2]. Batch normalization and ReLU activation function was used in all the convolution layers, except for the last one. where batch normalization was not used and the activation function was Tanh.
\begin{figure}[H]
  \centering
  \begin{minipage}[b]{1\textwidth}
    \includegraphics[scale=0.3]{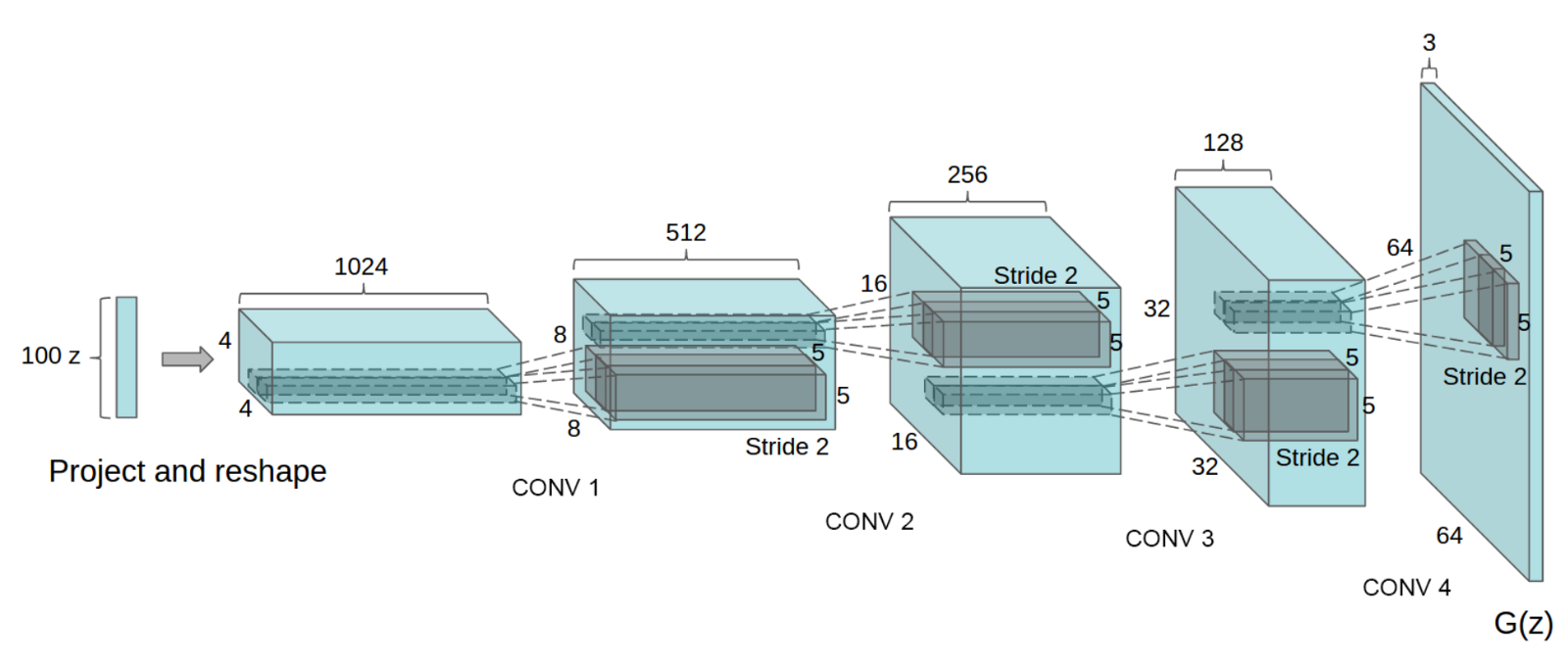}
	\caption{Generator of DCGAN: The input is a random noise with dimension 100, which is projected and reshaped in [batch\_size, 4, 4, 1024]. Then 4 convolution layers follow with kernel=[5, 5], stride=[2, 2] and output dimensions 512, 256, 128 and 3 respectively. The output is the fake image.}
  \end{minipage}
\end{figure}

\subsection{Discriminator}
\forceindent The discriminator works as a CNN, modified in order to avoid the GANs training difficulties. Having as inputs images of size 64x64, it outputs 3 predictions, which are valence, arousal and the real/fake prediction. In this case, four convolution layers are used with batch normalization following them. The activation function used is the leaky ReLU and dropout layers were used with dropout probability 0.5. After the last layer of convolution-dropout-batch norm-activation, Global Average Pooling was performed, which formed the final output of the discriminator. No activation function was necessary for the output, because this is a regression task, not a classification task. 

\capstartfalse
\begin{table}[H]
\centering
\begin{tabular}{|c|c|c|c|c|c|}
\hline
  Layer & filter & stride & padding & no of units\\
\hline
conv1 & [5, 5, 3, 64]   &[1, 2, 2, 1] & 'SAME'& \\
lrelu & &  & & \\
dropout &  & & & \\
\hline
conv2 & [5, 5, 64, 128]   &[1, 2, 2, 1] & 'SAME'& \\
batch norm &  & & & \\
lrelu & & &  & \\
dropout & &  & & \\
\hline
conv3 & [5, 5, 128, 256]   &[1, 2, 2, 1] & 'SAME'& \\
batch norm & & & & \\
lrelu & &  & & \\
dropout &  & & & \\
\hline
conv4 & [5, 5, 256, 512]  &[1, 2, 2, 1] & 'SAME'& \\
batch norm &  & & & \\
lrelu & &  & & \\
dropout &  & & & \\
\hline
GAP & &  & & \\
dense &  & & & 3\\
\hline
\end{tabular}
\end{table}
\capstarttrue
\begin{table}[H]
  \caption{Architecture of discriminator.}
\end{table}

\subsection{Loss}
\forceindent The loss for the Regression GAN has 2 parts, the discriminator and the generator loss. Those two losses are optimized individually and it is possible that they are updated different number of times during one iteration.

\forceindent For the discriminator part, the regression loss is first defined. As described in section 4.1, \\
\[L_{supervised} = 1 - ccc\] 
is the function that will be used, where CCC is the concordance coefficient correlation between the predicted and the real labels. However, the discriminator needs to be also evaluated on how good classified the real and the fake images. Hence the unsupervised loss will be(assuming that K+1 is the real/fake class):\\
\[L_{unsupervised}=-\{\mathbb{E}_{x\sim p_{data}(x)} \log \lbrack 1-p_{model}(y=K+1|x) \rbrack + \mathbb{E}_{x\sim G} \log \lbrack p_{model}(y=K+1|x) \rbrack \} \]

The total discriminator loss is defined as:
\[L_{discriminator} = L_{supervised} + L_{unsupervised}\]

\forceindent The generator loss also consists of two parts. First, it needs to be evaluated on how many of the samples that generated did the discriminator classified as real images.So, the loss for this part will be(assuming that K+1 is the real/fake class):
\[L_{G1}=-\mathbb{E}_{x\sim G} \log \lbrack 1-p_{model}(y=K+1|x) \rbrack\]
\forceindent The second part of the generator loss is called feature matching. As described in the Improved Techniques for GANs paper \cite{SSGAN}, feature matching is defined as “penalizing the mean absolute error between the average value of some set of features on the training data and the average values of that set of features on the generated samples”.

\forceindent Instead of mean absolute error, the Huber loss was used. The Huber loss \cite{Huber} for the difference of real and predicted data $a=y-f(x)$ is defined as:\\
\linebreak
\[L_\delta(y,f(x))=
\left \{
\begin{tabular}{ccc}
  $\dfrac{1}{2}(y-f(x))^2$ \quad \quad for $|y-f(x)| \leq \delta$ \\
  $\delta|y-f(x)|-\dfrac{1}{2} \delta^2$ \quad \quad otherwise 
  \end{tabular}
\right \}, \]
where $\delta$ is usually 1, y is the labels and f(x) the predicted values.
\linebreak
\linebreak
\forceindent Two of the most known loss functions are the Mean Absolute Error (MAE) and Mean Squared Error (MSE). The squared loss has the disadvantage that it has the tendency to be dominated by outliers values when the distribution is heavy tailed. As a solution to this problem, Huber loss combines much of the sensitivity of the MSE and the robustness of the MAE.

\begin{figure}[H]
  \centering
  \begin{minipage}[b]{1\textwidth}
    \includegraphics[scale=0.6]{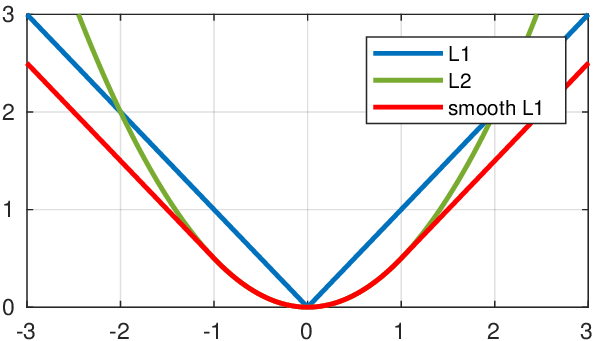}
	\caption{MAE vs MSE vs Huber (delta = 1.0) \\ \\ source: https://www.researchgate.net/figure/Plots-of-the-L1-L2-and-smooth-L1-loss-functions\_fig4\_321180616}
  \end{minipage}
\end{figure}

Hence, as defined above, the second part of the generator loss will be:\\
\[L_{G2} = \mathbb{E}_{x\sim p_{data}(x)} L_\delta(x,G(z)),\]
where x is the the real images and G(z) is the samples generated from noise z.

\forceindent Finally the total loss of the generator is:
\[L_G = L_{G1}+L_{G2}\]

\subsection{Training Tricks}
\forceindent Since GANs are difficult to train, some tricks had to be applied in the structure of the framework in order to avoid the fast convergence and stabilize training.
\begin{itemize}
  \item Higher learning rate is applied to the training of the generator.
  \item One-sided label smoothing is applied to the positive labels.
  \item Gradient clipping trick is applied 
  \item Reconstruction loss with an annealed weight is applied as an auxiliary loss to help the generator get rid of the initial local minimum.
\end{itemize}
\chapter{Technical Procedure}
\section{Pre-processing Scripts}
\forceindent Many procedures were needed in order to gather and get the data ready for the NN models. For every such need, python scripts were used. Most of the scripts were just executing data manipulation procedures with common libraries like Numpy. In case where a modification that concerns media streams (e.g. video) needed, it was implemented by using ffmpeg \cite{ffprobe} commands. But because these commands are executing in linux terminals for every video individually, an automated process had to be created. That was achieved by using the Subprocess module (Built-in module in python) for calling ffmpeg commands inside a loop, for all the videos.  
\section{TensorFlow}
\forceindent Tensorflow is an open source tool developed from Google and it is commonly used for machine learning and Deep learning applications. In the Tensorflow environment, the computations are performed with the use of data flow graphs. These graphs are are consisted of nodes and graph edges. The nodes represent mathematical operations and the edges are tensors which are passed between the nodes. Tensorflow offers its API in various programming languages, but the one used in this project is Python. Finally, CPU and GPU versions of the API are offered and since there was access to a GPU cluster by the university, the GPU version was used.

\section{Input Data}
\forceindent As mentioned in the Chapter 3, the final dataset is organized in 6 directories, 'frames'(big dataset), 'extra\_frames'(small dataset), 'test\_frames'(testing dataset) and 'labels', 'extra\_labels', 'test\_labels'. The frames directory contains one directory for every video, where the frames of this video are stored. The labels directory contains a text file for every video, where every line of a text is the valence and arousal values for the corresponding frame. Having the data in this form, the following process was followed in order to fit the data to the NN models.

\forceindent First, as stated in the dataset chapter, the size of all the videos that were used was 5 GB. So, since the videos had to be converted to raw images during the faces detection, it is obvious that the total size of all the images would be much bigger. More specifically, the size of the resulting dataset was 48 GB. Therefore, another way has to be used instead of loading the whole dataset into the machine, because it would obviously not fit in the machine's memory.

\forceindent In order to load the dataset into smaller pieces of data, the following was done. First, all the paths of all the images were loaded into a list. Then, some functions for loading and modifying the images were set. Finally, the data was loaded in mini batches from the paths list, following those load and modification functions for every such mini batch. This procedure was carried out by process\_data(data\_path, labels\_path, istraining) function, which is described in detail in the next section.

\section{Code Structure}
\forceindent The code that implements the basic functionality of our model is placed in just one file. This file is organized in methods. In this section, those methods are going to be described in detail:\\

\textbf{ huber\_loss(labels, predictions, delta=1.0):} Huber loss was described earlier as a loss function. This method takes as arguments the arrays with the predictions and the labels and also the delta value, which is by default 1. It executes all the appropriate calculations and it returns the Huber loss.
\\

\textbf{ lrelu(x, n, leak=0.2):} Leaky ReLU was described earlier as an activation function. This method takes as arguments the input x(ouput of batch normalization), the name of the variable n, and the slope, which is by default 0.2. It performs the activation and it give it as output.
\\

\textbf{ tf\_ccc(r1, r2):} Concordance coefficient correlation was also described earlier as a loss function. This method takes as input the predictions and the labels and returns the CCC as a float value. 
\\

\textbf{ process\_data(data\_path, labels\_path, istraining):} This method does all the processing that concerns the images and the labels. First it creates 3 lists. The first list is for the images paths, the second for the corresponding valence values and the third for the arousal values. Then, both lists are converted to tensors and are passed to input\_slice\_producer(), which create a queue for the data. The next step is to load images, reshape and rescale them as described in the previous chapter. Finally, the output batch is produced by train.batch() method and is returned.
\\

\textbf{ generator(input, random\_dim, is\_train, reuse=False):} This method contains the architecture for the generator network. The input is passed through all the layers described in the previous chapter and after the last activation function(Tanh), the method returns it as an output.
\\

\textbf{ discriminator(input, is\_train, reuse=False):} This method contains the architecture for the discriminator network. The input is passed through all the layers described in the previous chapter and after the Global Average Pooling, the method returns the logits as an output.
\\

\textbf{ train():} This is the main method that brings everything together and its functionality will be described briefly. First, the tensorflow's placeholders are created. Placeholders are just empty tensors with pre-defined dimensions that are fed later with data. Placeholders needed for this network are for the images(real\_image), the labels(labels), the random noise(random\_input) and a boolean which states if the feed is for training or testing(is\_train). Then, fake images is returned from the generator and both real and fake images are fed individually to the discriminator. Next step is the calculation of all the losses and updating variables. Since training and test data are needed, paths for both of them are defined and process\_data function is called twice, once for training and once for testing data. The session initialization follows. It is configured with GPU settings and it searches for any last checkpoint of the model. If it finds any, it loads the saved variables and it continues training, else it starts the training from scratch. The last part of this method is the loops of epochs and iterations, in which the data are fed to the model and the results are returned. Progress for training is saved for every iteration in a text file. Testing data are being fed to the model every 60 iterations and also new generated images are being saved every 60 iterations.
\section{Training}
\forceindent The optimization algorithm for the training of the model was the Adam optimizer with starting learning rate 0.0001 for the discriminator and 0.0002 for the generator. The discriminator maintains half of the learning rate of the generator, during the whole training process. Moreover, for the first GAN experiment generator and discriminator have the same update rate, which is one iteration each. For the second experiment, the update rate was 2 for the generator.
\section{Testing}
\forceindent Testing for this model include two parts, the generator and the supervised testing. Both parts are carried out every 60 iterations. For the supervised testing, \textbf{a batch of size 1000 from the testing dataset is fed} to the discriminator with boolean is\_train set to False. Then, the loss(1-CCC) is calculated and saved to a text file. For the generator testing, a random noise is fed to the generator and its output is returned. In order to evaluate the semi-supervised performance of the network, the artificial images are also fed to the discriminator in order to keep its output, which is the valence and arousal values. Since the output of the generator is the artificial images, they are saved in one image in a format of 8x8 (images number is same as the training batch size e.g. 64). Finally the text file with the output of the discriminator is saved with the same name of the image file.
\section{Evaluating Results}
\forceindent The results that concern the losses are evaluated by python scripts which process the text files created above. By doing that, there is freedom in design graphs and statistical values by using various python libraries.
\chapter{Evaluation of Results}
\section{CNN-M}
\forceindent As stated in Chapter 4, since the dataset is created from scratch, a first test has to be done, in order to check whether a model can actually learn from the data. So, the small dataset was used, in order to train a modified CNN-M network from \cite{kollias1,kollias2,kollias3}, which has already proved that it can return good results for the valence-arousal regression task.
\begin{figure}[H]
  \centering
  \begin{minipage}[b]{1\textwidth}
    \includegraphics[scale=0.77]{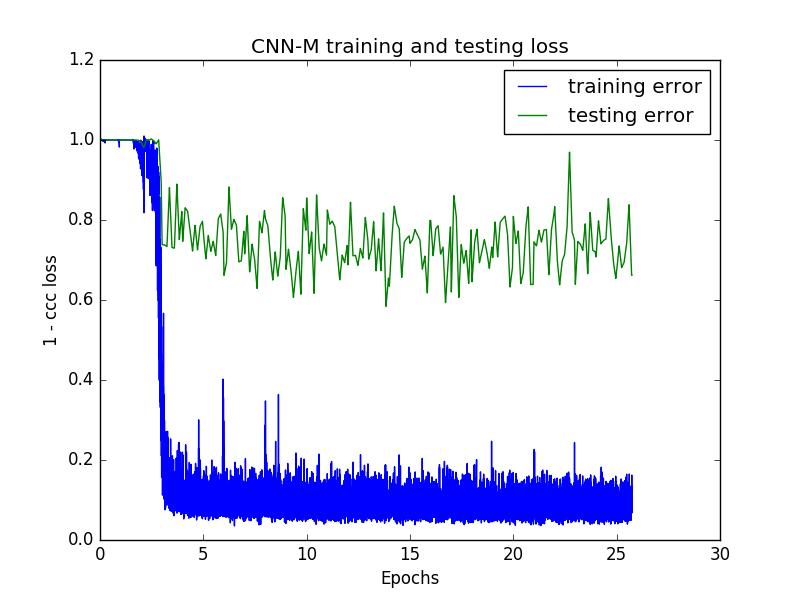}
	\caption{Loss of CNN-M with small dataset for training data and testing dataset.}
  \end{minipage}
\end{figure}

\capstartfalse
\begin{table}[H]
\centering
\begin{tabular}{|c|c|c|}
\hline
        &Training error(1-CCC) & Testing error(1-CCC)\\
        \hline
Minimum       & 0.04 & 0.58 \\
\hline
Mean         & 0.19      & 0.76\\
\hline
std deviation         & 0.28  &  0.1\\  
\hline
\end{tabular}
\end{table}
\capstarttrue
\begin{table}[H]
  \caption{Minimum, mean and standard deviation for training and testing error of CNN-M network.}
\end{table}
\forceindent By evaluating Figure 6.1, it can be concluded that the model actually learns from the data, because the training error converges near zero. However, there are some occasions that the graph does some spikes and it is not stable. This could happen either because of bad annotation, or from noise in the images. Nevertheless, the training error enables the use of other models with this dataset.

\forceindent Before looking to the above results, the model was expected to overfit and return a very bad testing error, due tot the small amount of data. Instead of this, there is a pretty good test error with minimum value 0.58. That means that the CCC between the real valence-arousal values and the predicted values is almost 0.4. According to the \cite{kollias1,kollias2,kollias3}, the CCC that was achieved from the same CNN-M network was 0.10-0.15, with a much bigger dataset than this. Two possible reasons for this to happen could be the following. First, that the valence-arousal distribution of the test dataset does not cover all the valence-arousal range of values and it returns good results for the same values. Second, because the annotation was carried out from one annotator it may has bias in some emotions and in combination with the small dataset, it returns good results.

\forceindent However, the testing error it is not very important in this case, because this model was used as a step that will confirm that the dataset is not useless.

\section{Regression GAN}
\subsection{Loss Evaluation}
\begin{figure}[H]
  \centering
  \begin{minipage}[b]{1\textwidth}
    \includegraphics[scale=0.9,left]{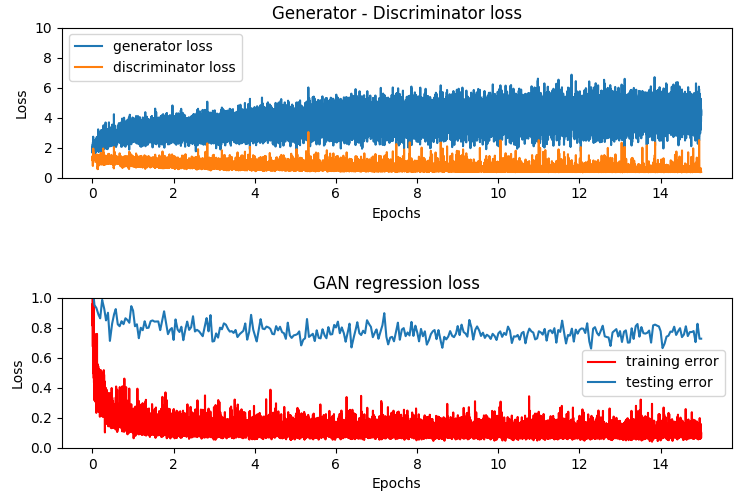}
	\caption{Loss of Regression GAN with small dataset for training and testing dataset.}
  \end{minipage}
\end{figure}

\capstartfalse
\begin{table}[H]
\centering
\begin{tabular}{lllllllllll}
       & Generator error & Discriminator error\\
Minimum       & 1.55 & 0.35 \\
Mean         & 3.7      & 0.68\\
std deviation         & 0.77  &  0.26\\  

\end{tabular}
\end{table}
\capstarttrue
\begin{table}[H]
  \caption{Minimum, mean and standard deviation for Generator and Discriminator error (trained with small dataset).}
\end{table}

\capstartfalse
\begin{table}[H]
\centering
\begin{tabular}{lllllllllll}
        &Training error & Testing error\\
Minimum       & 0.04 & 0.66 \\
Mean         & 0.12      & 0.77\\
std deviation         & 0.14  &  0.2\\  

\end{tabular}
\end{table}
\capstarttrue
\begin{table}[H]
  \caption{Minimum, mean and standard deviation for training and testing error(1-CCC) of Regression GAN network(trained with small dataset).}
\end{table}

\capstartfalse
\begin{table}[H]
\centering
\begin{tabular}{lllllllllll}
        &Discriminator accuracy \\
Real Images       & 0.99\\
Fake Images         & 0.07
\end{tabular}
\end{table}
\capstarttrue
\begin{table}[H]
  \caption{The discriminator accuracy after 18.000 iterations, for all the real images of the dataset and 12.000 fake images. }
\end{table}
\forceindent The above results are coming from the Regression GAN trained with the small dataset. Since it provides results that show that both unsupervised and supervised tasks are executed, the same model will be trained again with the big dataset. The only change in the training will be that the update rate of the generator will be set to 2.

\begin{figure}[H]
  \centering
  \begin{minipage}[b]{1\textwidth}
    \includegraphics[scale=0.9,left]{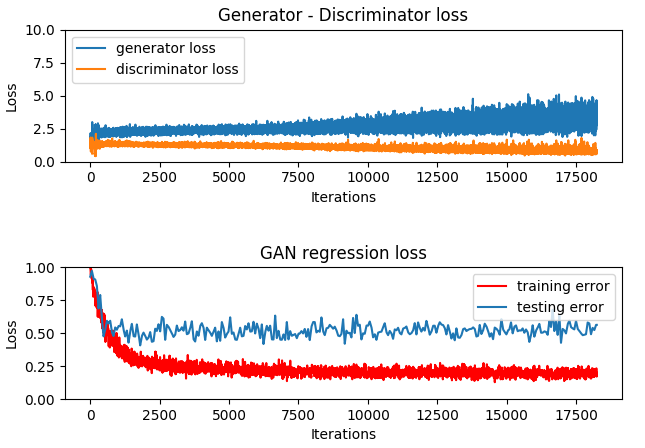}
	\caption{Loss of Regression GAN with big dataset for training and testing dataset .}
  \end{minipage}
\end{figure}
\capstartfalse
\begin{table}[H]
\centering
\begin{tabular}{lllllllllll}
       & Generator error & Discriminator error\\
Minimum       & 1.3 & 0.41 \\
Mean         & 2.66      & 1.06\\
std deviation         & 0.45  &  0.2\\  

\end{tabular}
\end{table}
\capstarttrue
\begin{table}[H]
  \caption{Minimum, mean and standard deviation for Generator and Discriminator error (trained with big dataset).}
\end{table}

\capstartfalse
\begin{table}[H]
\centering
\begin{tabular}{lllllllllll}
        &Training error & Testing error\\
Minimum       & 0.05 & 0.4 \\
Mean         & 0.24      & 0.52\\
std deviation         & 0.11  &  0.2\\  

\end{tabular}
\end{table}
\capstarttrue
\begin{table}[H]
  \caption{Minimum, mean and standard deviation for training and testing error(1-CCC) of Regression GAN network(trained with big dataset).}
\end{table}
\forceindent By observing the generator-discriminator loss(figure 6.3), it can be inferred that during the 18.000 iterations of training, the game between generator and discriminator is still played. The spikes in the loss graph is every time the generator fools the discriminator, then the discriminator improves itself and the generator tries to fool it again. However, the absolute value of the generator could be lower, in order to produce better fake images. 

\forceindent About the supervised loss (for the big dataset training), the training error(1-CCC) has a mean value of 0.24, which indicates that the model is getting trained both on predicting valence-arousal values and on predicting real-fake images. Moreover,  the testing error(1-CCC) has a mean value of 0.52. As mentioned in the previous case, this value is more than acceptable but it concerns only the current dataset, since the emotion annotation is a subjective procedure.

\forceindent In summary, it seems that the discriminator is able to find a way to split its weights in order to get trained for both the unsupervised and supervised tasks. Compared with the best results of the models implemented in \cite{kollias2}, this network performed very well, with always having in mind that the data are different in terms of annotation and because of that it may be easier to achieve good performance. It worths to remind that the CCC for the best model in \cite{kollias2} was 0.4, while for this model is 0.48, but that is just mentioned just for reference since different models trained with different data cannot be directly compared.
\capstartfalse
\begin{table}[H]
\centering
\begin{tabular}{|c|c|c|}
\hline
    Regression GAN    & Max CCC & Mean CCC\\
    \hline
 Small dataset      & 0.34 &  0.23\\
 \hline
 Big dataset         & 0.6  &   0.48  \\
 \hline
\end{tabular}
\end{table}
\capstarttrue
\begin{table}[H]
  \caption{Minimum and mean CCC of Regression GAN model for small and big database respectively.}
\end{table}
\subsection{Image Generation Evaluation}
\begin{figure}[H]
  \centering
  \begin{minipage}[b]{0.4\textwidth}
    \includegraphics[scale=0.3, left]{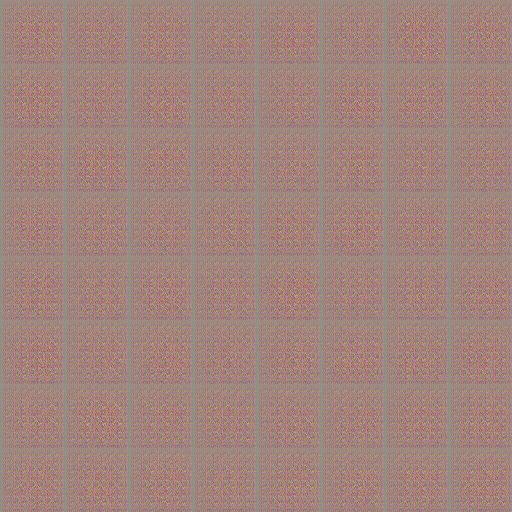}
    \caption{Fake images after 1 iteration.}
  \end{minipage}
  \hfill
  \begin{minipage}[b]{0.4\textwidth}
    \includegraphics[scale=0.3, left]{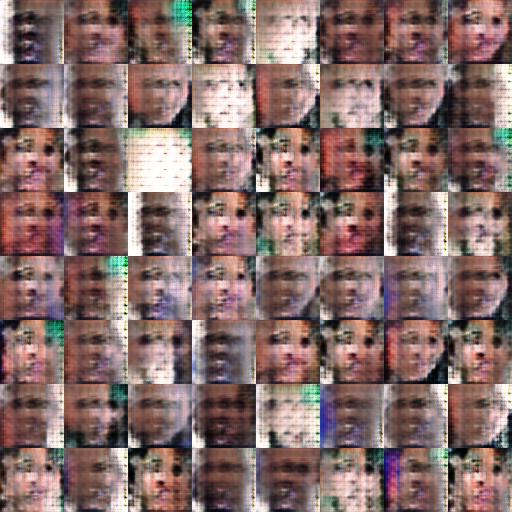}
    \caption{Fake images after 1.250 iterations.}
  \end{minipage}
\end{figure}

\begin{figure}[H]
  \centering
  \begin{minipage}[b]{0.4\textwidth}
    \includegraphics[scale=0.3, left]{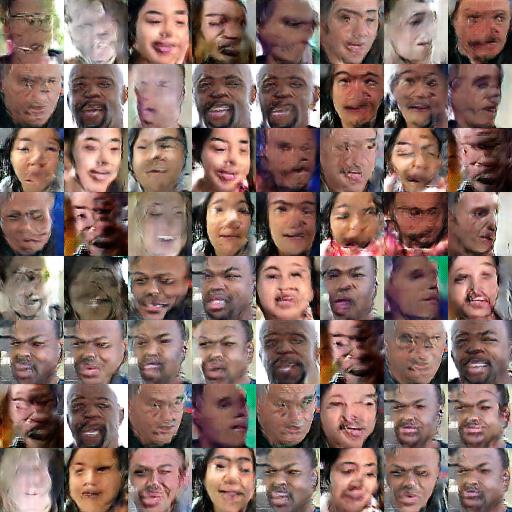}
    \caption{Fake images after 7.500 iterations.}
  \end{minipage}
  \hfill
  \begin{minipage}[b]{0.4\textwidth}
    \includegraphics[scale=0.3, left]{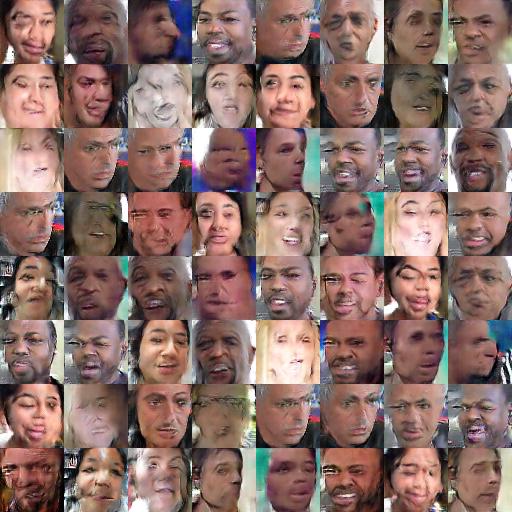}
    \caption{Fake images after 11.250 iterations.}
  \end{minipage}
\end{figure}

\begin{figure}[H]
  \centering
  \begin{minipage}[b]{0.4\textwidth}
    \includegraphics[scale=0.3, left]{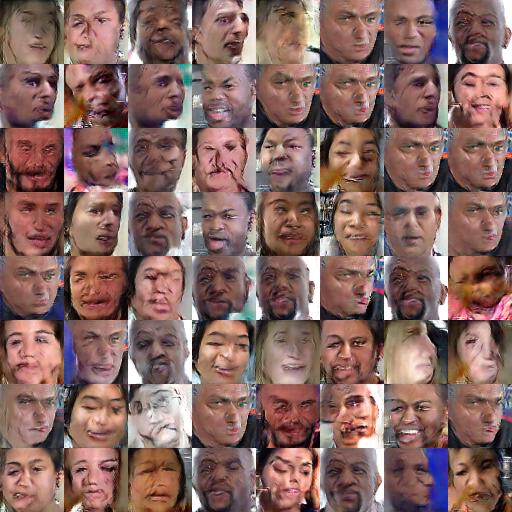}
    \caption{Fake images after 13.750 iterations.}
  \end{minipage}
  \hfill
  \begin{minipage}[b]{0.4\textwidth}
    \includegraphics[scale=0.3, left]{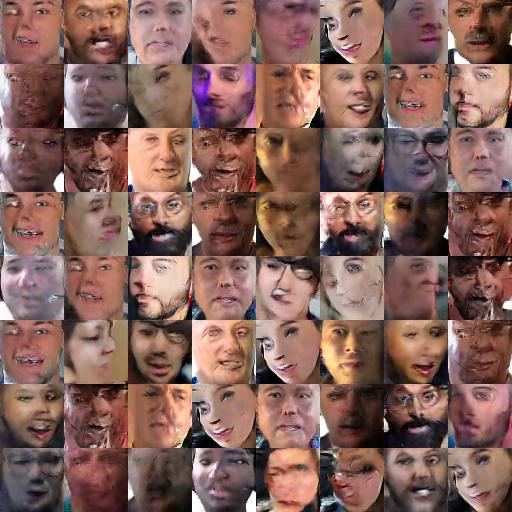}
    \caption{Fake images after 18.000 iterations.}
  \end{minipage}
\end{figure}

\forceindent In the above figures, the evolution of the generated images is shown. In figure 6.3 there is just the input noise and as the training continues, images are taking shapes that look like human faces. Moreover, there are features that get combined and make the faces more interesting(e.g. glasses, beard, etc). The quality of the images is not perfect and with more training or with more tuning of the model could be improved.

\subsection{Semi-Supervised Learning Evaluation}
\begin{figure}[H]
  \centering
  \begin{minipage}[b]{0.4\textwidth}
    \includegraphics[scale=0.45, left]{./figures/epoch1_iter2520}
    \caption{Fake images}
  \end{minipage}
  \hfill
  \begin{minipage}[b]{0.4\textwidth}
    \includegraphics[scale=0.45, left]{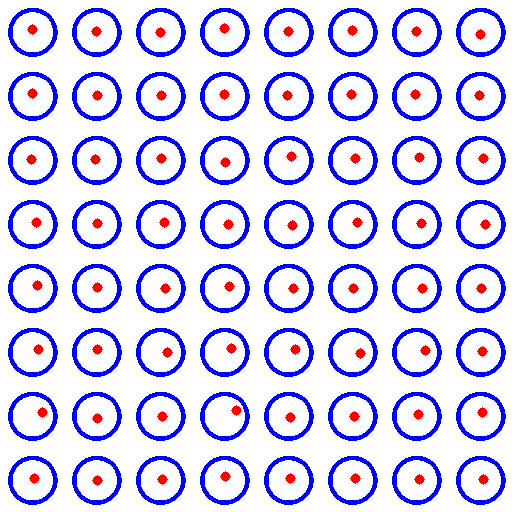}
    \caption{Valence-Arousal predictions for images.}
  \end{minipage}
\end{figure}

\begin{figure}[H]
  \centering
  \begin{minipage}[b]{0.4\textwidth}
    \includegraphics[scale=0.45, left]{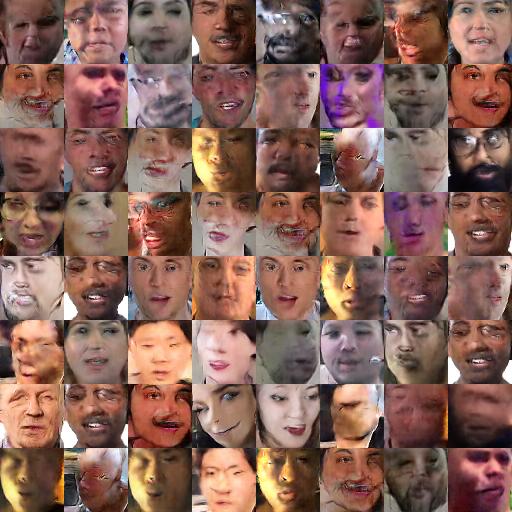}
    \caption{Fake images}
  \end{minipage}
  \hfill
  \begin{minipage}[b]{0.4\textwidth}
    \includegraphics[scale=0.45, left]{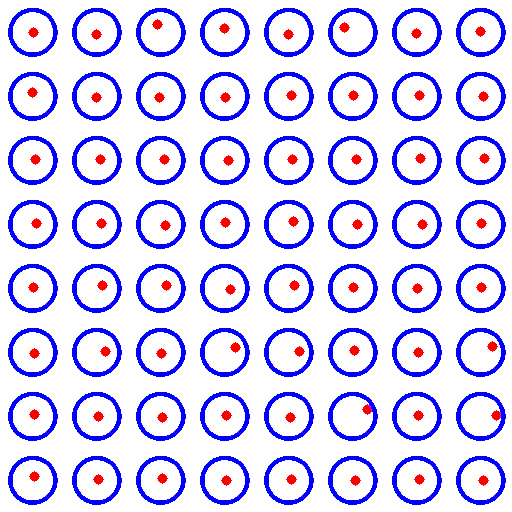}
    \caption{Valence-Arousal predictions for images.}
  \end{minipage}
\end{figure}

\begin{figure}[H]
  \centering
  \begin{minipage}[b]{0.4\textwidth}
    \includegraphics[scale=0.45, left]{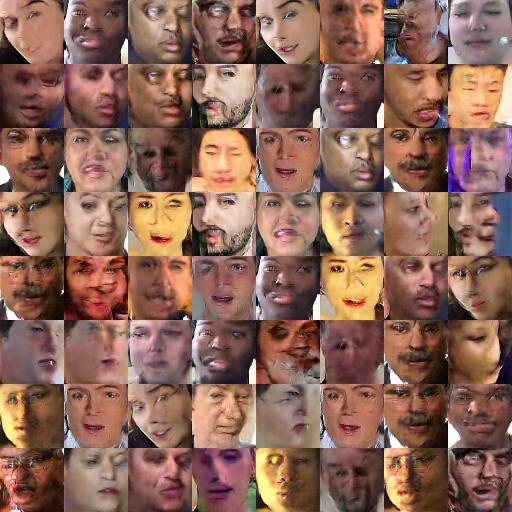}
    \caption{Fake images}
  \end{minipage}
  \hfill
  \begin{minipage}[b]{0.4\textwidth}
    \includegraphics[scale=0.45, left]{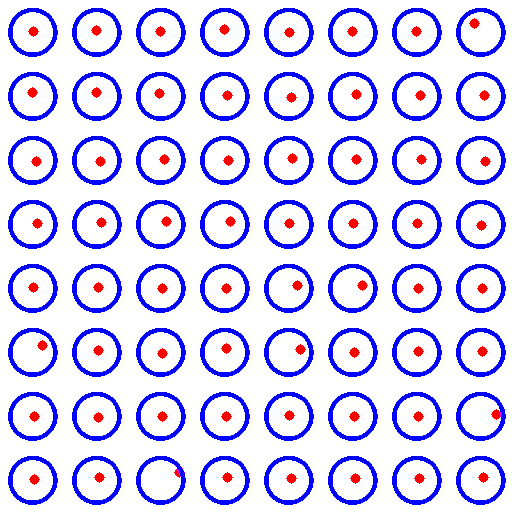}
    \caption{Valence-Arousal predictions for images.}
  \end{minipage}
\end{figure}

\begin{figure}[H]
  \centering
  \begin{minipage}[b]{0.4\textwidth}
    \includegraphics[scale=0.45, left]{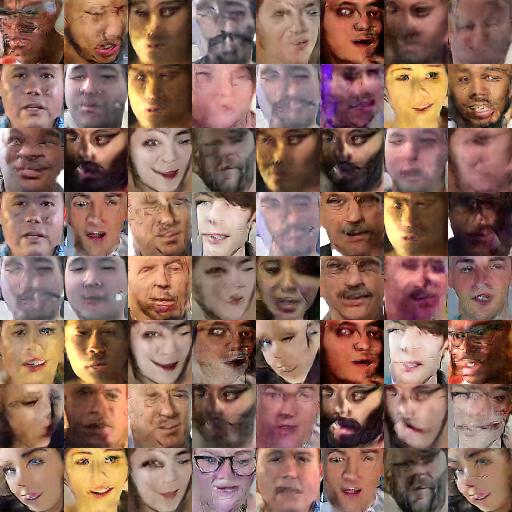}
    \caption{Fake images}
  \end{minipage}
  \hfill
  \begin{minipage}[b]{0.4\textwidth}
    \includegraphics[scale=0.45, left]{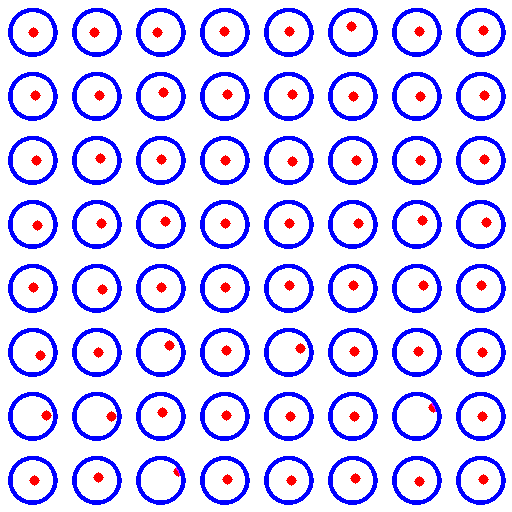}
    \caption{Valence-Arousal predictions for images.}
  \end{minipage}
\end{figure}

\forceindent The figures above (6.9 - 6.16) are very interesting on how the semi-supervised learning performs. The evaluation is done just by observing the images and the predicted values side by side. The first thing that can be observed from these figures is that the predictor avoids big absolute values of arousal and that there are no many negative valence values. However, one can say that this reflects how the annotations of the training data is distributed, as shown in Chapter 3.

\forceindent Some images where a face cannot be displayed clearly is not easy to evaluate the predictions results. For the images that it is not hard to tell that there is a face, there is a good number of occasions where the predictions seem to agree with the reality. It is certain that there is a big space for improvement in order to have acceptable emotion predictions in generated samples.
\chapter{Conclusion and Future Work}
\section{Dataset Summary}
\forceindent The dataset created in this project could be proved useful in many machine learning projects \cite{kollias4,goudelis2013exploring,kollias5,doulamis1999interactive,kollias15}, because it follows the "in the wild" concept. This concept is very useful in the real world, because any applications that analyze human behavior, such as emotions recognition, should be able to be used in real situations. If a model is trained with data recorded inside a laboratory, then when is applied in the real world it will probably be inaccurate, due to the factor of people's unpredictability, which cannot not been taken in account.

\forceindent The dataset quality is high in terms of the images it includes. The face detector was very accurate and there is no much noise \cite{raftopoulos2018beneficial} that can hurt any machine learning model. In most cases, faces are clear and the image is cut without enabling something other than the face being in the image. The detector had no issue to also detect faces with glasses or with the hand in front of the mouth and other situations like these. This also helped to support the "in the wild" concept, as mentioned before. However the dataset can be enriched with more unique people, in order to enrich the diversity of the data and with higher absolute values of valence and arousal, in order to maintain a good emotion distribution.  

\forceindent As mentioned before, the annotation task was carried out from one person. That said, the labels in many cases may be biased and not close the the proper value that a psychologist would have suggested. For the original "aff-wild database" \cite{kollias1,kollias2,kollias3} the annotation was much more objective, since it was carried out by 6-8 annotators and more specialized techniques were used in order to ensure annotation quality.  
\section{Neural Networks Summary}
\forceindent After creating a database and before trying complex deep learning models, it was decided to use an already implemented CNN network, in order to test the dataset. For this reason, the modified CNN-M from \cite{kollias1,kollias2,kollias3} was chosen. There was no need to use a deeper network like ResNet L50 or VGG, because the purpose was only to overfit the data and then continue with the main model of the project. Indeed, the outcome of the model was a very low training error and surprisingly a very good testing error, in compare with the CNN results that were presented in Chapter 2. However, as mentioned earlier, there were reasons like annotation bias that could have returned better results than the reality.

\forceindent When the background and related work was studied, it seemed like a good idea to combine two successful concepts like emotions predictions and images generation. First, from previous works, it was proved that CNNs had a decent performance with the "in the wild" data. Second, from the background study, GANs had a good performance on generating images, especially with faces. So, the idea was to implement a combination of emotions predicting and images generation.

\forceindent Finally a regression GAN was implemented. The results presented in the previous chapter indicate that it can work and produce decent results. But since GANs are difficult to train, a reliable and stable model needs more time to be implemented. However, this type of network can be really useful for data augmentation tasks, if someone takes advantage of the semi-supervised attribute of the model.  
\section{Future Work}
\forceindent Emotions recognition "in the wild" with valance and arousal values is a very promising concept, that has already produced very good results in many papers. In this project, it was attempted to combine GANs with the prediction procedure by implementing a Regression GAN in a newly created dataset. Of course, there is a lot of work that could be done in order to achieve good and reliable results that could compete other state of the art implementations. 

\forceindent The dataset created for this project having as guidance the "aff-wild" dataset could be enriched with more data. First, in order to help GANs produce better fake images, the dataset should be enriched in terms of diversity, with more different and unique people. Second, the data should have more distributed valence and arousal values especially in very low or high values. Then, the data would really help the unsupervised and the supervised part of the Regression GAN to perform better.

\forceindent There is a lot of space for tuning in the project's network. However, something that may produce better results, would be the use of Capsule GANs. As described in Chapter 2, Capsule GANs have the ability not to only learn the features of the image, but they also keep their spatial information. This is achieved by substituting the discriminator of a GAN, which is actually a CNN version, with a capsule network. It was shown in \cite{CapsuleGANS} that they can provide better results than the traditional GANs.

\bibliography{refs}
\bibliographystyle{ieeetr}
\begin{appendices}
\chapter{LSEPI Checklist}
\includegraphics[scale =0.7]{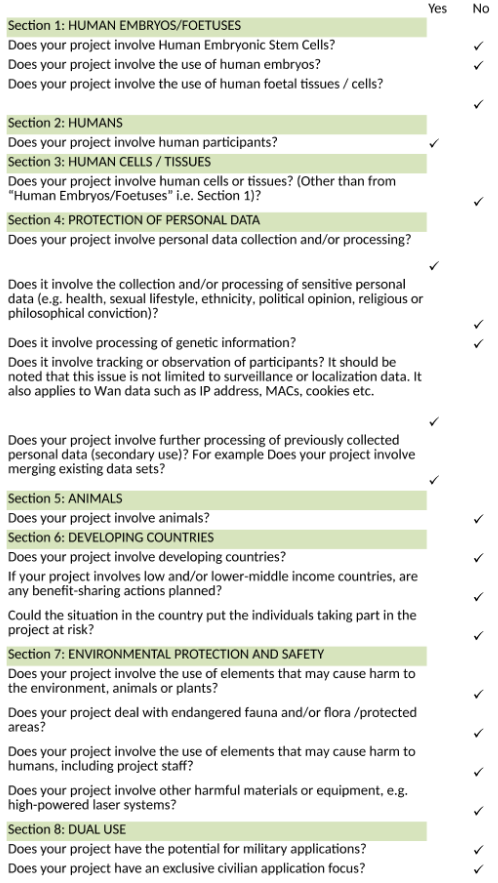}
\newpage
\includegraphics[scale =0.85]{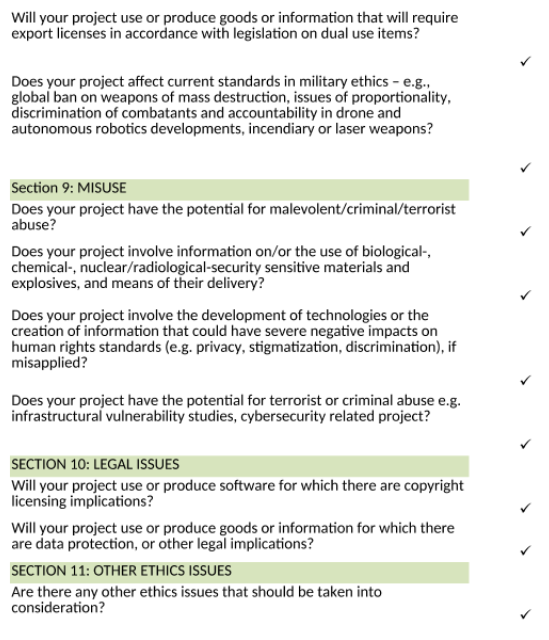}\\
\forceindent When the word "data" is used in this project, it is referred to images which contain human faces. The expressions of the faces in the images, which come from online sources, are analyzed by deep learning models, in order to predict the emotions. So, the project does contain humans, 68 males and 38 females, which in majority they are adults. These images were generated from videos that were collected online. As mentioned before, the emotions predictions are based on the observation of peoples expressions in those images. Finally, some already existing databases were presented in Chapter 2 but none of them was merged with the one that was created during the project.

\end{appendices}

\end{document}